\documentclass{CUP-JNL-CHR}

\usepackage{graphicx}
\usepackage{multicol,multirow}
\usepackage{amsmath,amssymb,amsfonts}
\usepackage{rotating}
\usepackage{appendix}
\usepackage[sort&compress]{natbib}
\usepackage[colorlinks,allcolors=blue]{hyperref}
\usepackage{breakurl}
\usepackage{wrapfig}
\usepackage[export]{adjustbox}

\articletype{Research Article}
\jname{}
\jyear{2025}
\jdoi{}

\begin{document}

\begin{Frontmatter}

\title{\textit{Looking for the Inner Music}: Probing LLMs' Understanding of Literary Style}

\author*[1]{Rebecca M.\ M.\ Hicke}
\author[2]{David Mimno}

\authormark{Hicke and Mimno}

\address[1]{\orgdiv{Department of Computer Science}, \orgname{Cornell University}, \orgaddress{\city{Ithaca}, \state{NY}, \country{USA}}}

\address[2]{\orgdiv{Department of Information Science}, \orgname{Cornell University}, \orgaddress{\city{Ithaca}, \state{NY}, \country{USA}}}



\corresp{\email{rmh327@cornell.edu}}



\abstract{Recent work has demonstrated that language models can be trained to identify the author of much shorter literary passages than has been thought feasible for traditional stylometry.
We replicate these results for authorship and extend them to a new dataset measuring novel genre.
We find that LLMs are able to distinguish authorship and genre, but they do so in different ways.
Some models seem to rely more on memorization, while others benefit more from training to learn author/genre characteristics.
We then use three methods to probe one high-performing LLM for features that define style.
These include direct syntactic ablations to input text as well as two methods that look at model internals.
We find that authorial style is easier to define than genre-level style and is more impacted by minor syntactic decisions and contextual word usage. However, some traits like pronoun usage and word order prove significant for defining both kinds of literary style.}


\end{Frontmatter}

\section*{Plain Language Summary}

Recent work has demonstrated that language models can be trained to identify the author of much shorter literary passages than has been thought feasible for traditional stylometry.
We replicate these results for authorship and extend them to a new dataset measuring novel genre.
These results offer an intriguing possibility: rather than just identifying a minimal set of characteristics that \textit{distinguish} author styles, can we identify a maximal suite of characteristics that \textit{define} an author's or genre's style?
We begin with a series of black-box classification experiments.
These show that LLMs are able to distinguish authorship and genre, but also that different model families appear to balance style identification and source memorization differently.
We then use three methods to probe one high-performing LLM for features that define style.
These include direct syntactic ablations to input text as well as well as probes into cross-attention values and contextual word embeddings.
We find that authorial style is easier to define than genre-level style and is more impacted by minor syntactic decisions and contextual word usage. However, some traits like pronoun usage and word order prove significant for defining both kinds of literary style.

\section{Introduction}

Measuring and characterizing literary style has long been a fascinating but difficult problem in computational literary studies.
Lexical methods have frequently been successful in distinguishing authorial signatures, but are often less compelling as support for a literary theory.
For example, Hamilton's use of \textit{upon} may be sufficient to distinguish his contributions to the Federalist Papers from Madison's \citep{mosteller1972deciding}, but it is not a particularly exciting result from a stylistic perspective.
In addition, these lexical features do not allow us to determine what, if any, stylistic signals exist in very short text segments.
Literary style at this small scale has rarely been explored using computational methods, largely because until recently there were few clear ways to do so.
New AI technologies have demonstrated the ability to identify subtle patterns in complex text.
Paradoxically, however, their power and flexibility can make explaining \textit{how} they make decisions difficult.
Explaining how models are able to identify style could be valuable for NLP research as well as literary studies: if we are able to leverage these technologies to identify and characterize literary style, we may be able to provide stylistically relevant insights into the features that characterize language usage.
In addition, by applying these new technologies to a highly semantically and syntactically complex literary problem, we can glean insight into their linguistic abilities.

In this work, we evaluate the ability of contemporary large language models (LLMs) to distinguish both author and genre in very short (20--50 word) passages.
We find that all tested models and baselines are able to recognize author and genre in these texts with above random accuracy, confirming that stylistic signals do exist at this scale. The largest LLMs --- a quantized Llama-3 8b \citep{llama3} and Flan-T5 Xl \citep{chung2022scaling} --- achieve the highest performance on both tasks, with over 50\% accuracy at attributing texts to one of 27 authors and over 70\% accuracy attributing texts to one of five genres. Further exploration into these models' behavior suggests that the Llama-3 model makes use of information memorized during pre-training, whereas the Flan-T5 model builds representations of authorial and genre-level style throughout fine-tuning.

Based on these results, we then seek to find interpretable features that characterize the model's internal representations.
We pursue this goal in two ways. First we modify the texts in the test dataset and evaluate the models on these perturbed texts. Second, we analyze the model's internal representations.
Both of these experiments reveal that authorial and genre-level style are distinguished by different features and that authorial style is easier to characterize. Minor syntactic elements, such as punctuation and capitalization patterns, appear to contribute solely to authorial style. In addition, contextual word usage seems to be highly significant for authorial style, but its impact on genre style is less noticeable. We hypothesize that many of the features that easily identify authorial styles are clouded in genre styles because many authors with their own authorial styles contribute to each genre. Genre-level style may be characterized by broader topical trends instead of fine-grained linguistic features.

Despite the differences between authorial and genre-level style, some trends apply to both. Although previous research back to \citet{mosteller1972deciding} has tended to deliberately ignore less frequent content terms, we find that stop words are not more important for characterizing style than content words in either task; in fact, it seems that both frequently and infrequently used words are used to characterize style. Of stop words, pronouns appear to be the most important and they play an important role in distinguishing the language usage of many authors and genres. Finally, we see that word order is extremely important for models' ability to identify style across both tasks, implying that contextual language models are finding sequence-level information not carried by lexical information alone.
While the question of characterizing literary style will take considerable effort beyond this collection and work, these results offer an intriguing first step in moving beyond simple classification accuracy.

\section{Related Work}

\subsection{Authorship Attribution}

Considerable previous research has evaluated computational approaches to authorship attribution \citep{huang2024authorship}. Recently, many studies in this area have focused on distinguishing between human and LLM-generated texts; however, for this work our interest lies solely in distinguishing between texts written by human authors. Many ``lexical, syntactic, semantic, structural, and content-specific'' features have been used in computational authorship attribution approaches, including ``character and word frequencies, parts of speech, punctuation, topics, and vocabulary richness'' \citep{huang2024authorship}. These features are often used as inputs for a variety of statistical and machine learning methods including support vector machines (SVMs), logistic regression, recurrent neural networks (RNNs), long short-term memory (LSTMs), Siamese networks, and more \citep{huang2024authorship}. 

Since the introduction of transformer-based large language models (LLMs) in 2017 \citep{vaswani2017attention}, scholars have also explored their use for authorship attribution. BERT and its many variants have been applied to this task, often in combination with techniques like contrastive learning, gradual unfreezing, and slanted triangular learning \citep{huang2024authorship}. Recent work has also explored the ability of larger encoder-decoder \citep{hicke2023t5, najafi2022text} and decoder-only \citep{huang2024alms, huang2024large, adewumi2024limitations, wen2024aidbench} models to perform attribution.

However, of these studies only \citet{hicke2023t5} and \citet{adewumi2024limitations} use literary data in their analyses. In addition, only \citet{huang2024large} examine what features models use to attribute texts, and they do so only by asking the models to self-report their reasoning. In contrast to these works, we do not study authorship attribution with the intention of creating or evaluating maximally effective attribution methods. Instead, our interest is in studying what signals of literary style exist at very small scales and to what extent LLMs can help us identify and analyze these signals. To this end, we focus on what LLM performance can tell us about literary style and about LLMs' capabilities, including a series of ablation and probing experiments into what features the models use for attribution.

\subsection{Genre Identification}

Genre identification is often a more difficult to define than authorship attribution due to the complexity of genre as a concept; unlike authorship, variable criteria may be used to define a genre and what texts belong to it \citep{underwood2016genres}. However, despite these challenges, many computational approaches to genre identification have been proposed and evaluated. Some use bag-of-word features along with statistical and light-weight machine learning methods like logistic regression, Naive Bayes, SVMs, and more \citep{allison2011quantitative, underwood2016genres, underwood2013mapping, hettinger2016classification, sharmaa2020rise}. Other works use similar statistical and ML methods along with additional features like emotion arcs \citep{kim-etal-2017-investigating} or more in-depth linguistic annotations \citep{calvo2021novel}. Yet another branch of research on genre identification uses unsupervised clustering algorithms with bag-of-words features \citep{al2018stylometric}, social network features \citep{ardanuy2015clustering, coll-ardanuy-sporleder-2014-structure}, or embeddings and topic classifications \citep{sobchuk2024computational}.

Recent research has also explored the effectivity of transformer-based LLMs for genre identification. Much of this research has focused on non-literary definitions of genre \citep{UCHIDA2024100089, kuzman2023troubling, roussinov2025controlling, make5030059}; these studies have largely found model performance promising, although they have also identified some weaknesses, such as expanding to cross-domain data. Two studies, \citet{liu2020deepgenre} and \citet{bamman2024classification}, have explored using a range of LLMs for literary genre identification. Both studies look at $\sim$500 word or token chunks of literary texts. \citet{liu2020deepgenre} evaluate several statistical methods, including a Naive Bayes algorithm, and a fine-tuned BERT model for detecting the genre of paragraphs from novels that have a large proportion of genre-relevant keywords. \citet{bamman2024classification} similarly evaluate linear and logistic regression, fine-tuned BERT, RoBERTa, and Llama-3 8b models, and few-shot prompted GPT-4o, Llama-3 70b, and Mixtral 8x22b models for detecting the genre of randomly selected novel passages. \citet{liu2020deepgenre} find that BERT performs the best out of all tested methods and \citet{bamman2024classification} find that the Llama-3 8b, GPT-4o, and Llama-3 70b models do best. Again, these results suggest that LLMs show promise for genre identification tasks. However, no study except \citet{bamman2024classification} explores what features the models use to perform genre identification, and even \citet{bamman2024classification} only ask the chat-enabled models self-report on distinguishing genre characteristics.

Again, our work differs from those cited because its primary interest is in using LLM behavior to evaluate what signals of genre-level style exist in very short text segments, and not optimizing the model's performance on this task.

\section{Data}

We use two datasets throughout this work: one for authorship attribution and one for genre identification. The authorship attribution dataset is based on a pre-existing dataset\footnote{\url{https://github.com/computationalstylistics/100_english_novels}} compiled by the Computational Stylistics Group\footnote{\url{https://computationalstylistics.github.io}} which contains 100 novels written by 33 authors in English in the 19th and early 20th centuries. To process this dataset for our use, we standardize punctuation using regular expressions and strip all front and back matter, chapter headings, footnotes, and formatting marks from each text file by hand. We then split each file with the \texttt{nltk} sentence tokenizer, keeping only sentences between 20 and 50 words long. Finally, we only retain works by authors for whom all three novels have at least 675 sentences of sufficient length. The final corpus thus contains 81 novels by 27 authors written between 1839--1937. The complete list of authors and novels included in this dataset can be found in Appendix \ref{sec:authorCorpusAppendix}.

\begin{wrapfigure}{r}{0.5\linewidth}
  \FIG{
  \includegraphics[width=0.95\linewidth]{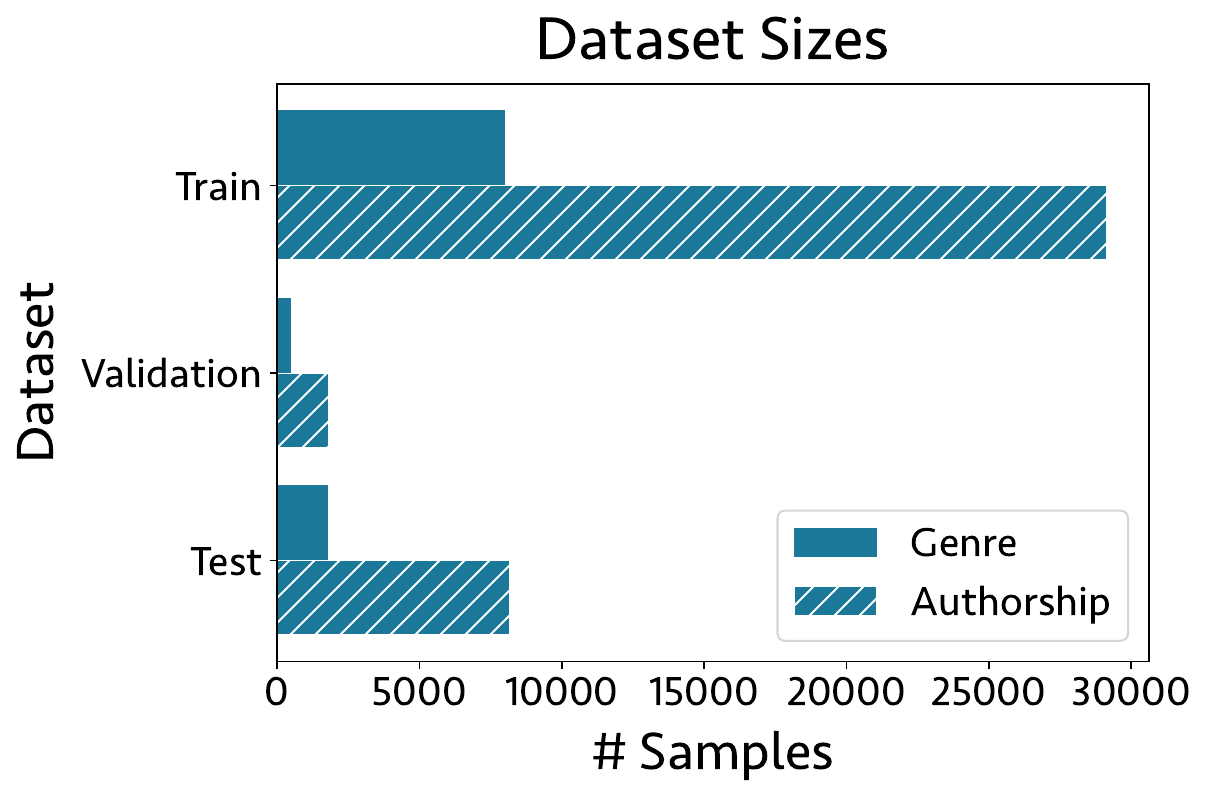}}
  {\caption{The size of each dataset used in the experiment by number of samples.}
  \label{fig:datasetSizes}}
\end{wrapfigure}

In addition to the existing authorship attribution dataset, we introduce a new genre-identification dataset. It contains six books each from five genres: fantasy, historical fiction, horror, mystery, and science fiction. To select the texts, we first sort the relevant Project Gutenberg\footnote{\url{https://www.gutenberg.org}} genre tag by number of downloads. Then, we include each successive novel in the dataset only if no book already in the dataset is by the same author, the genre of interest is listed as one of the first two genre tags on Goodreads,\footnote{For historical fiction, the genre tag can be prefaced by `classics' and `fiction'.} and no other genre of interest is included in the first two tags on Goodreads. We continue this process until five novels are selected for each genre. We then use the same pre-processing and tokenization pipeline on these texts\footnote{The text file versions of each novel we use are drawn from Project Gutenberg.} as we did for the authorship attribution dataset. Each processed text has at least 400 sentences between 20 and 50 words long. The final dataset contains 30 novels by 30 authors, all written between 1726 and 2003. The full list of authors and novels in this dataset can be found in Appendix \ref{sec:genreCorpusAppendix}.

To distinguish memorization of training data from true task performance, we divide the full datasets into training and testing subsets. We also create small validation sets for model selection.
The test sets contain excerpts from one completely unseen novel for each author or genre, along with unseen portions of novels included in the training and validation sets.
Specifically, for each of the 54 novels in the authorship attribution dataset that are not withheld from training, we include 540 randomly sampled sentences in the training dataset, 34 sentences in the validation dataset, and 101 sentences in the test dataset. 101 randomly sampled sentences from each withheld novel are also included in the test dataset. Similarly, from each of the 25 non-withheld novels in the genre identification dataset, we include 320 randomly sampled sentences in the training dataset, 20 sentences in the validation dataset, and 60 sentences in the test dataset. 60 sentences from each withheld novel are also added to the test dataset. The final number of samples in each dataset is displayed in Figure \ref{fig:datasetSizes}. We withhold a novel from each class of interest in order to distinguish between a model's ability to identify the characteristics of a \textit{novel} (via the partially held-out works) and the characteristics of a \textit{genre or author} (via the fully held-out works).

\section{Classification}

Although our ultimate goal is to characterize literary style, we begin by establishing that information exists in short passages that can distinguish between authorial and genre styles, and that LLMs are able to use this information.
While this phase resembles a standard NLP evaluation, we emphasize that its purpose is not to find the overall best-performing model. This phase \textit{is not intended to  be an exhaustive evaluation of LLM-based literary style identification}, but rather to establish that literary style exists and is observable under these conditions.

We compare two LLM families and two baseline methods on both literary style identification tasks: authorial attribution and genre identification. Following \citet{hicke2023t5}, we select models from the T5 family; however, in contrast to the cited paper we focus on the instruction-tuned Flan-T5 models \citep{chung2022scaling} to allow for zero-shot model probing. Specifically, we evaluate the small, base, large, and xl Flan-T5 models. We also evaluate a more recent, decoder-only model --- Llama-3 with 8 billion parameters \citep{llama3} --- which we fine-tune using 4-bit quantization\footnote{\url{https://unsloth.ai}} and LoRA \citep{hu2022lora}. We fine-tune the Flan-T5 and Llama-3 models to produce class labels (author or genre names) as free-text output using the formats in Table \ref{exampleInputOutputs}. We then compare these models to the two bag-of-words baselines: a support vector machine (SVM) with a linear kernel and TF-IDF unigram values as features (the highest performing baseline from \citet{hicke2023t5}) and cosine delta, an established authorship attribution method \citep{smith2011improving}. Implementation details for each method are provided in Appendix \ref{sec:parameterAppendix}.

\begin{table}[t]
\TBL{\caption{Example formatted input and output pairs for each model type. For genre identification, replace ``AUTHOR'' with  ``GENRE.'' \texttt{<extra\_id\_0>} is the masking tag used during the masked language modeling pre-training for the T5-family models.}\label{exampleInputOutputs}}{\begin{fntable}\tabcolsep=2pt
\begin{tabular}{lll}
\toprule
\textbf{Model} & \textbf{Example \:Input} & \textbf{Example Output}\\
\midrule
Flan-T5 & AUTHOR: \texttt{<extra\_id\_0>} | This is example text. & AUTHOR: John Doe | This is example text. \\
\botrule
Llama-3 & This is example text. | AUTHOR: & This is example text. | AUTHOR: John Doe \\
\botrule
\end{tabular}
\end{fntable}}
\end{table}

\subsection{Comparing Models}

\begin{figure}[ht]
  \FIG{
      \includegraphics[width=0.4\textwidth]{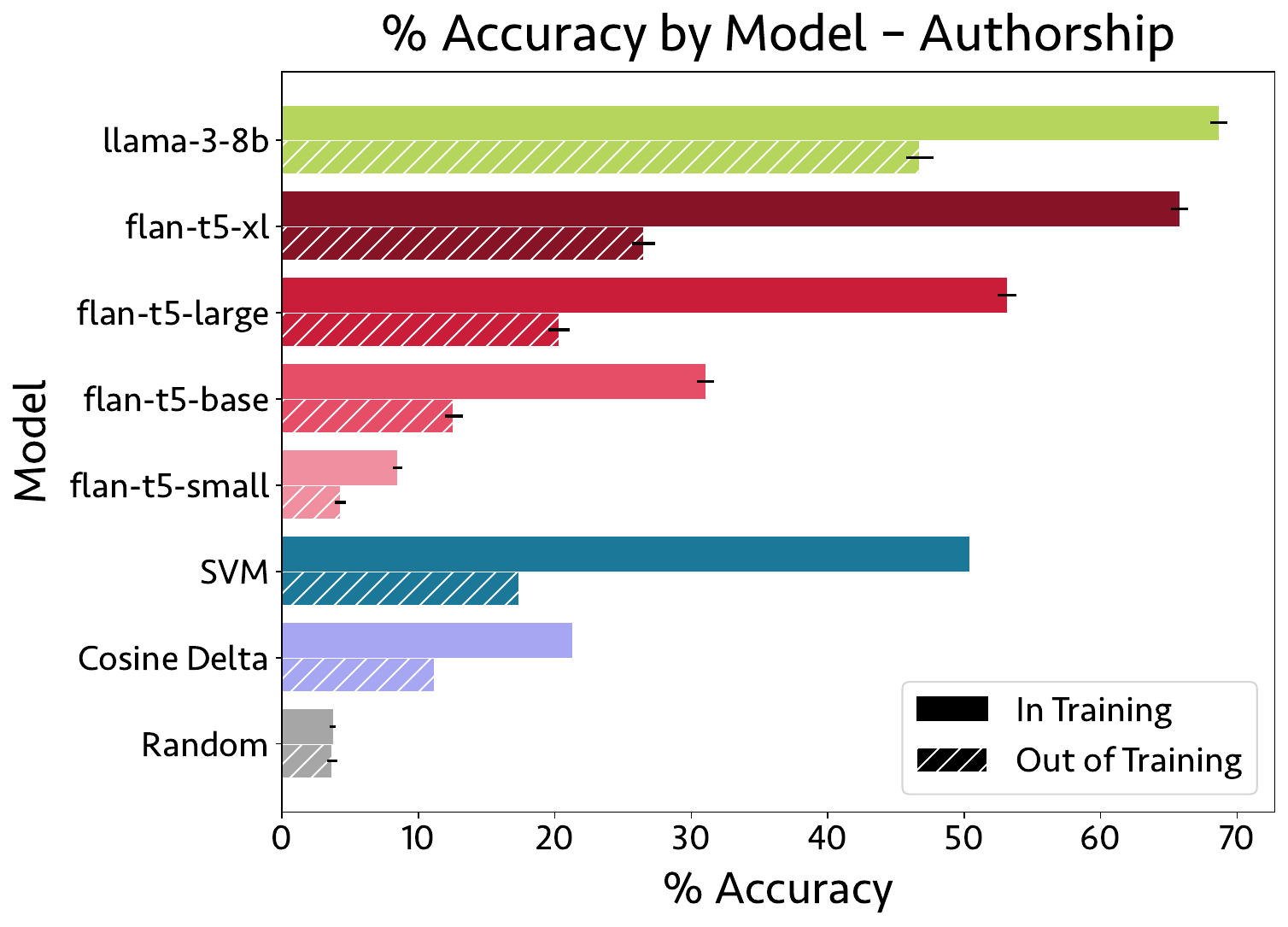} \hspace{6ex}
      \includegraphics[width=0.4\textwidth]{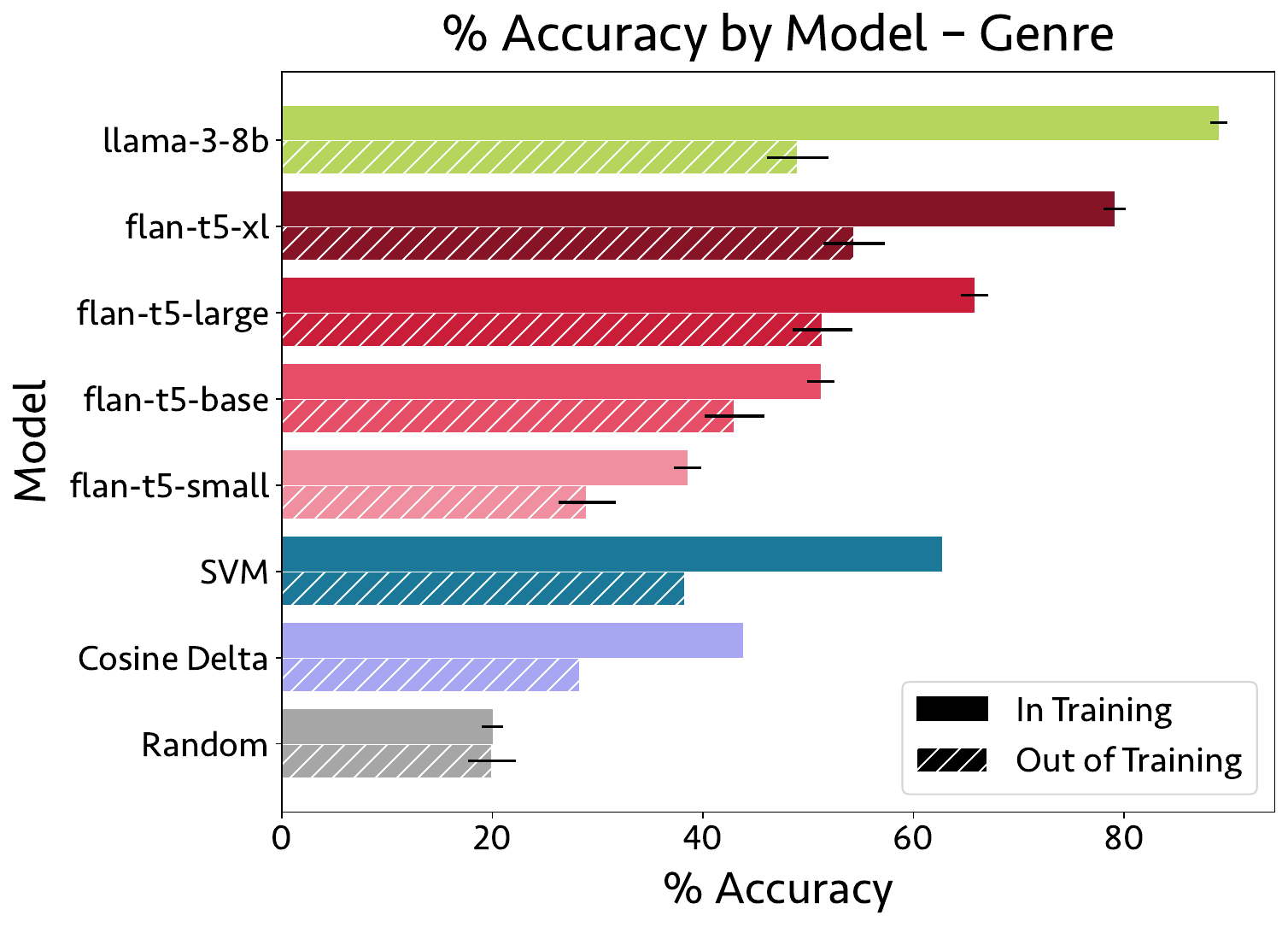}}
  \caption{The overall accuracy of each model for (left) authorship attribution and (right) genre identification. Accuracy is separated into results for samples from novels included in training and samples from novels withheld from training. Results of a single run are reported and error bars represent the standard error bootstrapped over 1000 iterations. The y-axis is sorted by model's performance on samples from in-training novel.}
  \label{fig:overallAccuracy}
\end{figure}

All models and baselines achieve above random performance for authorship attribution and genre identification (Figure \ref{fig:overallAccuracy}). This confirms both that signals of literary style exist in 20--50 word texts and that all tested methods are capable of recognizing these signals. In addition, all methods achieve much higher accuracy for genre identification than authorship attribution; this may suggest that genre is easier to identify in short texts, or may result from the difference in the number of possible classes between the datasets (5 vs.\ 27). 

Unsurprisingly, all methods perform better on excerpts from novels partially included in training than excerpts from novels fully withheld from training (Figure \ref{fig:overallAccuracy}).
This difference implies two things:
that the results for pre-trained models are not relying purely on memorization from pre-training, and 
that the stylistic signals used by each method are partially novel-specific. However, the two largest and best performing models --- \texttt{llama-3-8b} and \texttt{flan-t5-xl} --- still outperform the random baselines by $\sim 2.5\times$ for genre identification and $\sim6.5\times$ for authorship identification on samples from withheld novels, indicating that the models learn some generalizable stylistic traits.

These large generative models outperform all other methods (Figure \ref{fig:overallAccuracy}), with \texttt{llama-3-8b} achieving the highest overall accuracy on both tasks. While both baselines --- the SVM and cosine delta --- achieve impressive performance on both identification tasks, they are outstripped by the larger transformer-based models. For these transformer-based LLMs, we see that model performance increases with size, although the performance increase per parameter decreases as size grows. 

\subsection{Probing Memorization}

\begin{figure}[t]
  \FIG{
  \includegraphics[width=0.7\linewidth]{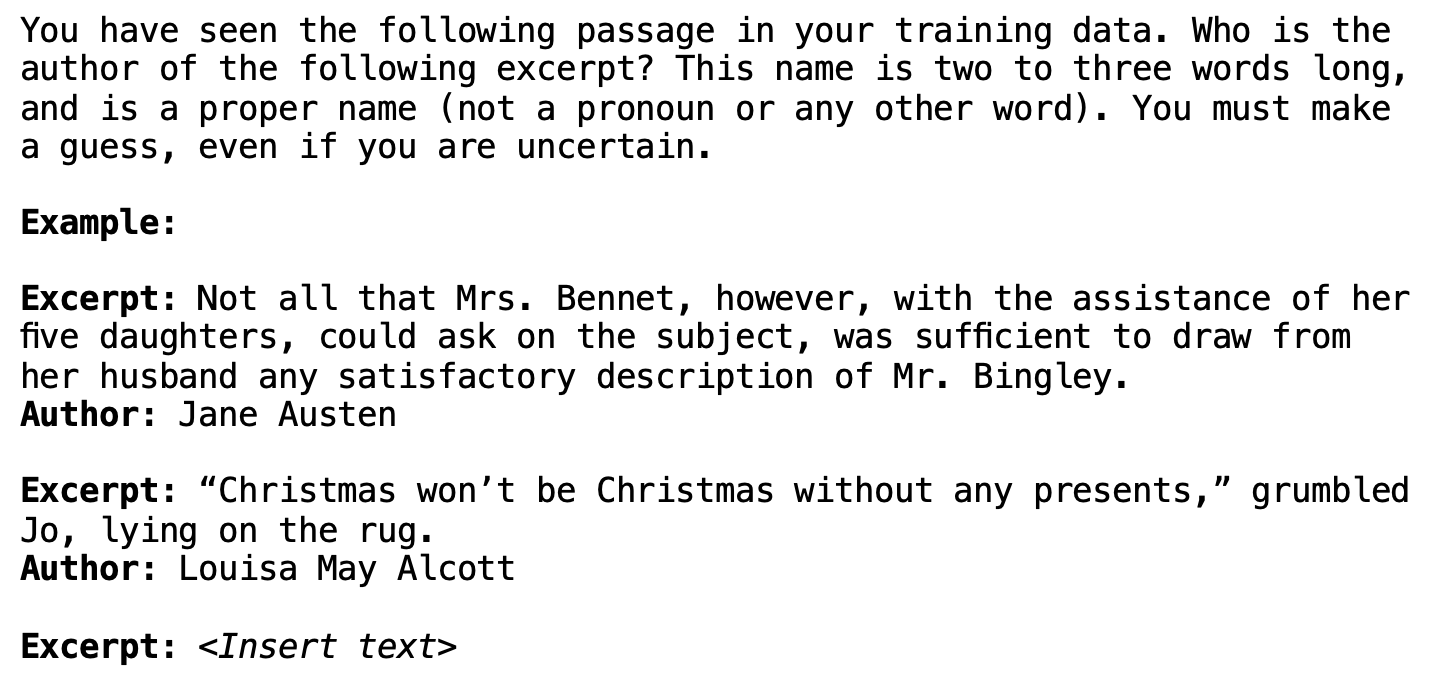}}
  {\caption{The prompt used to probe \texttt{llama-3-8b} and \texttt{flan-t5-xl} for memorization of studied texts.}
  \label{fig:authorPrompt}}
\end{figure}

The high performance of the larger LMs may be due to memorization of the texts and associated information during pre-training. To test whether \texttt{llama-3-8b} and \texttt{flan-t5-xl} are using memorized information, we adapt the prompt from \citet{chang-etal-2023-speak} (Figure \ref{fig:authorPrompt}). We then use this prompt to query the base version of both models for the author of every excerpt in the authorship attribution and genre identification datasets and record a match if the model's unconstrained generation matches the correct author name. We use authorship attribution to probe for memorization in both datasets because the idea of genre is more generalizable than that of author. The only ways for a model to correctly report the author of a text are a) if it has memorized the text and author or b) if it is able to correctly match the text to a stylistic profile of an author. We assume that the base versions of these models are unlikely to have a stylistic profile of each author in our dataset, and therefore assume that they will only return the correct author for a high proportion of text segments if they have memorized the texts. In contrast, it seems more likely that the models have developed stylistic conceptualizations of genre, and therefore we deem probing for this trait a less effective method of testing for memorization. 

We find that \texttt{llama-3-8b} correctly provides the author of 33.11\% (SE: 0.51) of excerpts from the authorship attribution dataset and 42.06\% (SE: 1.13) of samples from the genre identification dataset, suggesting this model has memorized many of the included texts. For samples from novels withheld from training, fine-tuning \texttt{llama-3-8b} only leads to a 13.31\% increase in attribution accuracy. In addition, of the 5,021 excerpts correctly attributed by the fine-tuned model from the authorship attribution dataset, 2,177 (43.36\%) are also correctly attributed by the prompted model. Surprisingly, however, the accuracies by author of the prompted \texttt{llama-3-8b} are uncorrelated with the accuracies of the fine-tuned model ($p=0.2$). For example, the prompted model correctly identifies the most excerpts by Charles Dickens (88.45\%), but the fine-tuned model's performance on Dickens is among its worst (44.22\%); in fact, of the 134 samples correctly identified by the fine-tuned model, 133 are also correctly identified by the prompted model. This may suggest that, throughout fine-tuning, some of the information previously memorized by the model is lost (thus the drop in accuracy for Charles Dickens) and that capacity is instead used to store information on the authors for whom no information was memorized (so the accuracy for Marie Corelli goes from 1.3\% to 62.05\% with fine-tuning).

To determine whether \texttt{llama-3-8b}'s memorization of texts is correlated to author popularity, we borrow a proxy measure of author popularity --- the length (in characters) of their Wikipedia page --- from \citet{d2023chatbot}. We find that there is, indeed, a strong positive correlation between the popularity of an author and the prompted model's accuracy at attributing excerpts from that author for both the authorship attribution dataset (Pearson's R: 0.65, $p < 5 x 10^{-4}$) and the genre identification dataset (Pearson's R: 0.60, $p < 5\times10^{-4}$). However, no significant correlation exists between author popularity and the fine-tuned model's performance on the authorship attribution dataset ($p = 0.75$) or author popularity and the fine-tuned model's ability to provide the correct genre of an excerpt by that author ($p=0.56$). Furthermore, we find a correlation between author popularity and the proportion of excerpts correctly attributed by the fine-tuned \texttt{llama-3-8b} model that are also correctly attributed the prompted model (Pearson's R: 0.58, $p < 5\times10^{-3}$). Thus, it appears that the fine-tuned model may be relying on memorized information when attributing quotes from popular authors and information learned through fine-tuning when attributing quotes from less well-known writers.


In contrast to \texttt{llama-3-8b}, \texttt{flan-t5-xl} only correctly provides the author of one sample from the authorship test dataset (0.01\%, SE: 0.01) and four samples from the genre test dataset (0.22\%, SE: 0.01). These correctly attributed excerpts are all from relatively popular authors, but so few exist that no conclusive insights can be drawn from these results. As with \texttt{llama-3-8b}, there is no significant correlation between the popularity of authors in the authorship attribution dataset and the fine-tuned \texttt{flan-t5-xl} model's accuracy at attributing excerpts from these authors ($p=0.71$) or the fine-tuned \texttt{flan-t5-xl} model's ability to provide the correct genre of excerpts by an author ($p = 0.95$). 

Because of the stark differences between the behavior of \texttt{llama-3-8b} and \texttt{flan-t5-xl}, we hypothesize that they use different information to perform style identification: while \texttt{llama-3-8b} has access to memorized information from pre-training which it may be leveraging, \texttt{flan-t5-xl} appears to be learning stylistic representations from fine-tuning.

\begin{figure}[t]
  \FIG{
  \includegraphics[width=0.7\linewidth]{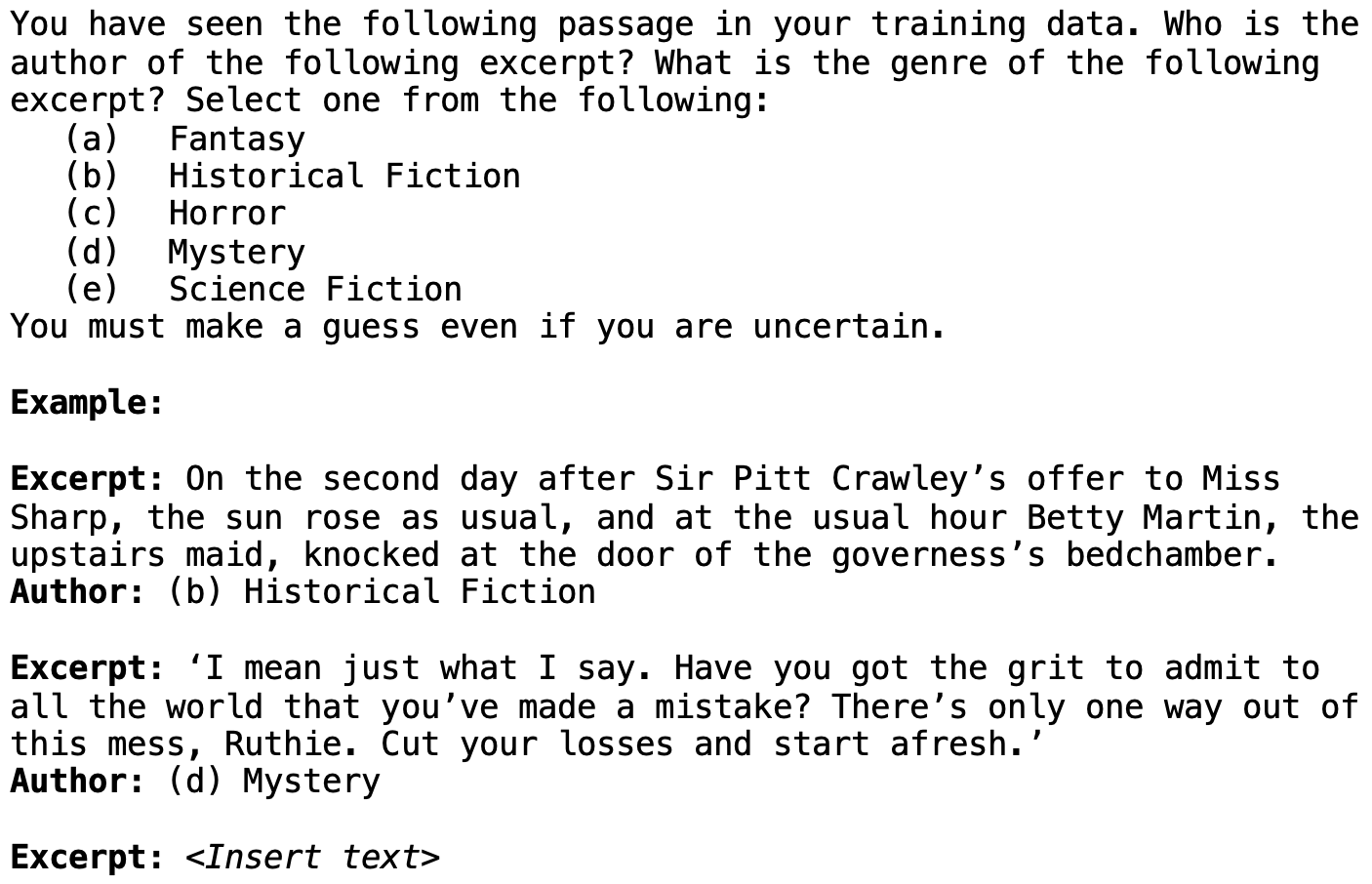}}
  {\caption{The prompt used to probe \texttt{llama-3-8b} and \texttt{flan-t5-xl} for internal representations of genre.}
  \label{fig:genrePrompt}}
\end{figure}

We further evaluate whether the base \texttt{llama-3-8b} and \texttt{flan-t5-xl} models are able to predict the genre of excerpts from the genre identification test dataset. While we take accurately reporting the author of an excerpt to indicate memorization of text, this is not necessarily true for genre; models may have an internal representation of genre that can be applied without fine-tuning. For this probe, we again use a prompt adapted from \citet{chang-etal-2023-speak} (Figure \ref{fig:genrePrompt}). In this prompt, we include a list of possible genres since it is difficult to define a ``correct'' genre in an unconstrained setting. We find that \texttt{llama-3-8b} and \texttt{flan-t5-xl} are able to correctly report the genre of 56.89\% (SE: 1.18) and 44.39\% (SE: 1.17) of excerpts respectively, suggesting that these models do have internal representations of genre that can be leveraged.

\begin{figure}[t]
  \FIG{
  \includegraphics[width=0.9\linewidth]{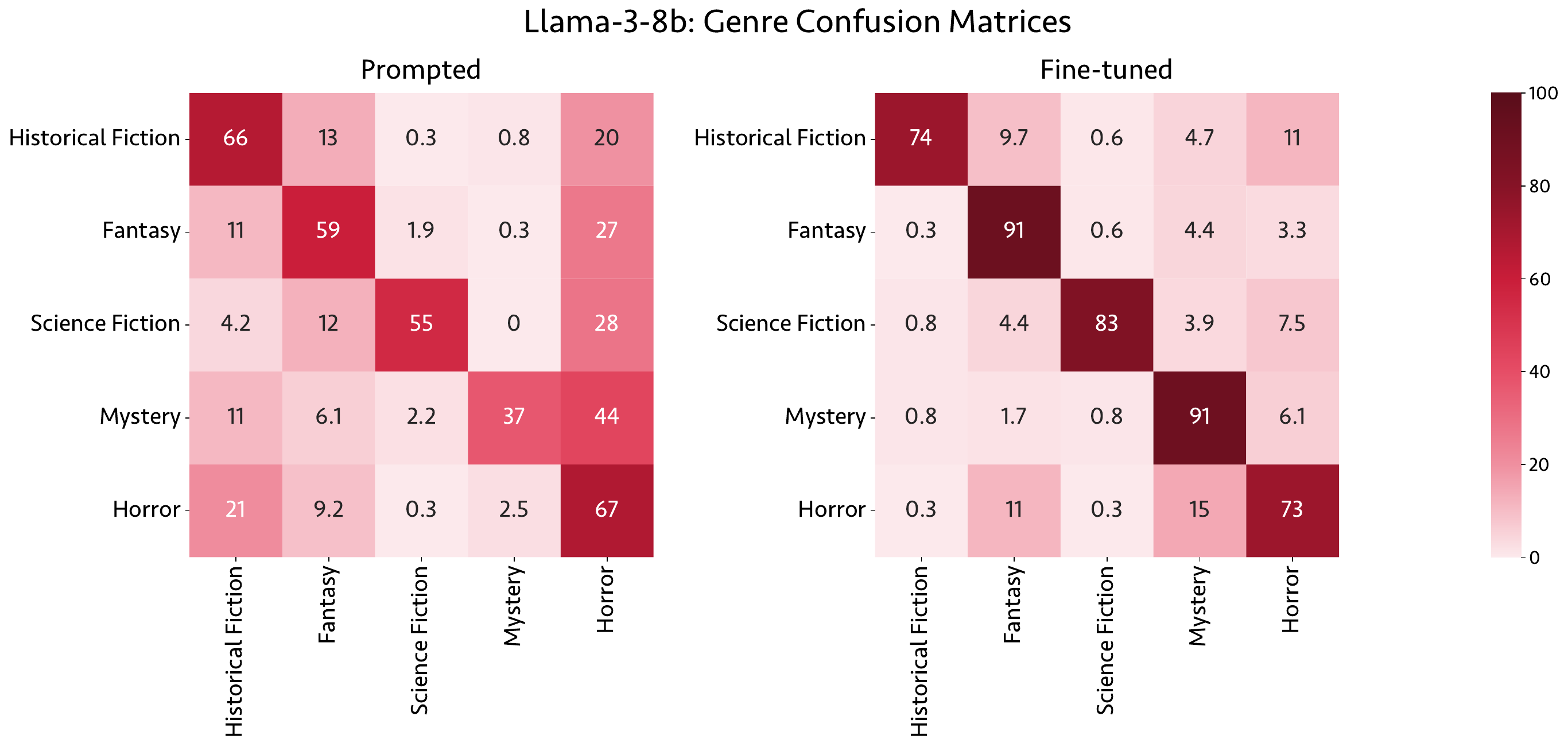}}
  {\caption{Confusion matrices of the responses of the prompted (left) and fine-tuned (right) \texttt{llama-3-8b} models for genre identification. Correct labels are represented by rows and model outputs are columns; labels produced outside of the correct set are ignored. The rows sum up to $\sim$100\%.}
  \label{fig:llamaGenreConfMatrix}}
\end{figure}

\begin{figure}[t]
  \FIG{
  \includegraphics[width=0.9\linewidth]{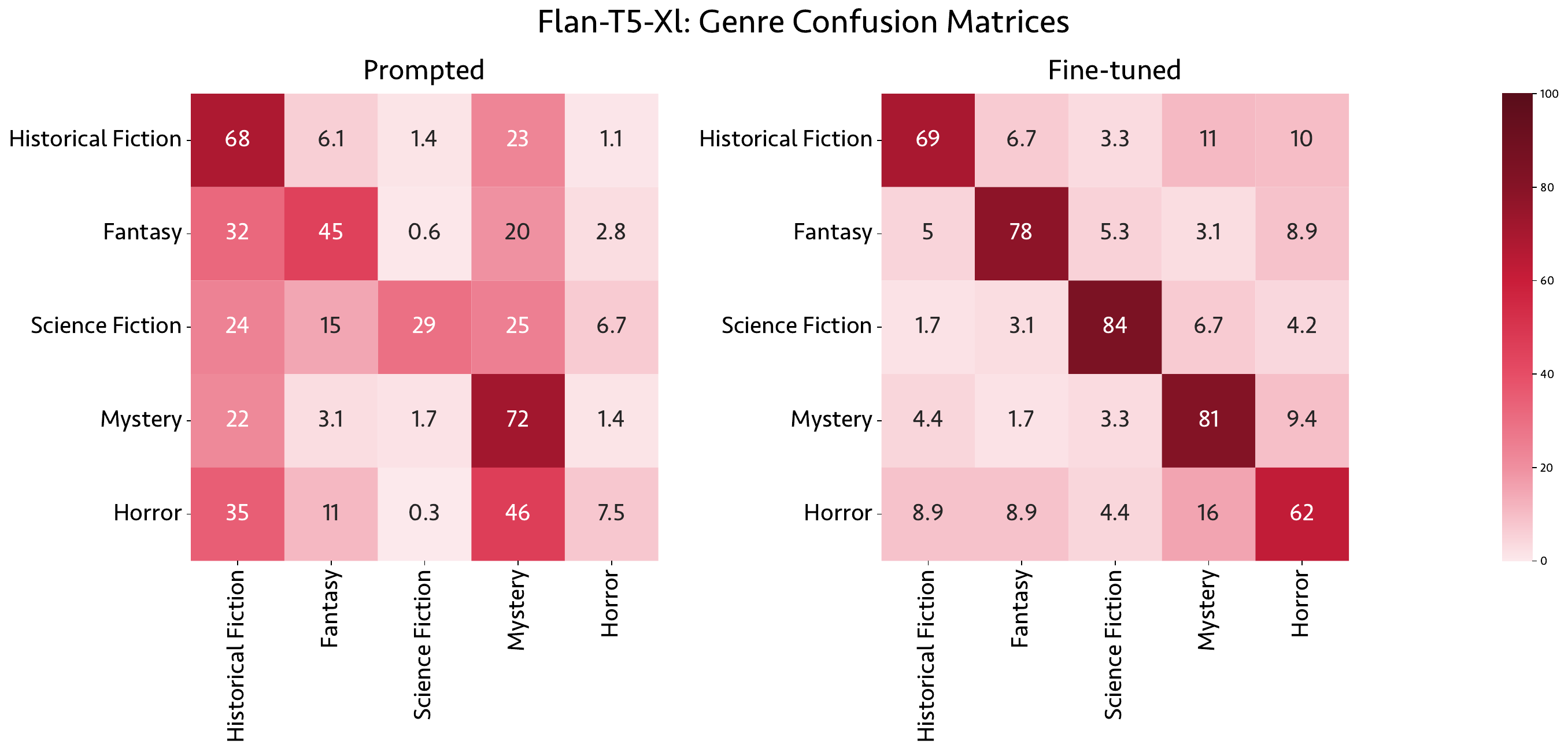}}
  {\caption{Confusion matrices of the responses of the prompted (left) and fine-tuned (right) \texttt{flan-t5-xl} models for genre identification. Correct labels are represented by rows and model outputs are columns. The rows sum up to $\sim$100\%.}
  \label{fig:t5GenreConfMatrix}}
\end{figure}

There are not significant correlations between the accuracy per genre of the fine-tuned and prompted \texttt{llama-3-8b} models ($p=0.16$) or \texttt{flan-t5-xl} models ($p=0.54$). If we examine the confusion matrices of responses from the fine-tuned and prompted versions of both models (Figures \ref{fig:llamaGenreConfMatrix} and \ref{fig:t5GenreConfMatrix}), we see that both prompted models tend to incorrectly label excerpts as historical fiction or fantasy, while the fine-tuned models do not. The prompted \texttt{llama-3-8b} model also has a strong tendency to mislabel excerpts as horror whereas the prompted \texttt{flan-t5-xl} model tends to mislabel excerpts as mystery. Both fine-tuned models also mislabel some excerpts as horror and mystery, although the extent to which these labels are assigned to misattributed samples does not follow the same pattern as the prompted models. Interestingly, science fiction is the label to which the least samples are miss-assigned in almost all cases, suggesting there is something unique about the genre. Thus, overall, we see that there are some trends in how the models treat genre, although these trends are not broadly generalizable.

Because the prompted \texttt{flan-t5-xl} model could not provide the authors of these texts, it seems unlikely that the genres of the identified excerpts were memorized at the book level; instead, we hypothesize that the model is able to recognize signals of genre-level style without fine-tuning. This may contribute to the models' much higher performance with genre identification than authorship attribution. Overall, fine-tuning only improves \texttt{flan-t5-xl}'s accuracy on out-of-training excerpts by 14\% and only improves \texttt{llama-3-8b}'s by 3.67\%. Again, this suggests that \texttt{flan-t5-xl} is `learning' more from fine-tuning than quantized \texttt{llama-3-8b}, although both models have a leveragable internal representation of genre before fine-tuning.

\begin{figure}[ht]
  \FIG{
      \includegraphics[width=0.4\textwidth, valign=t]{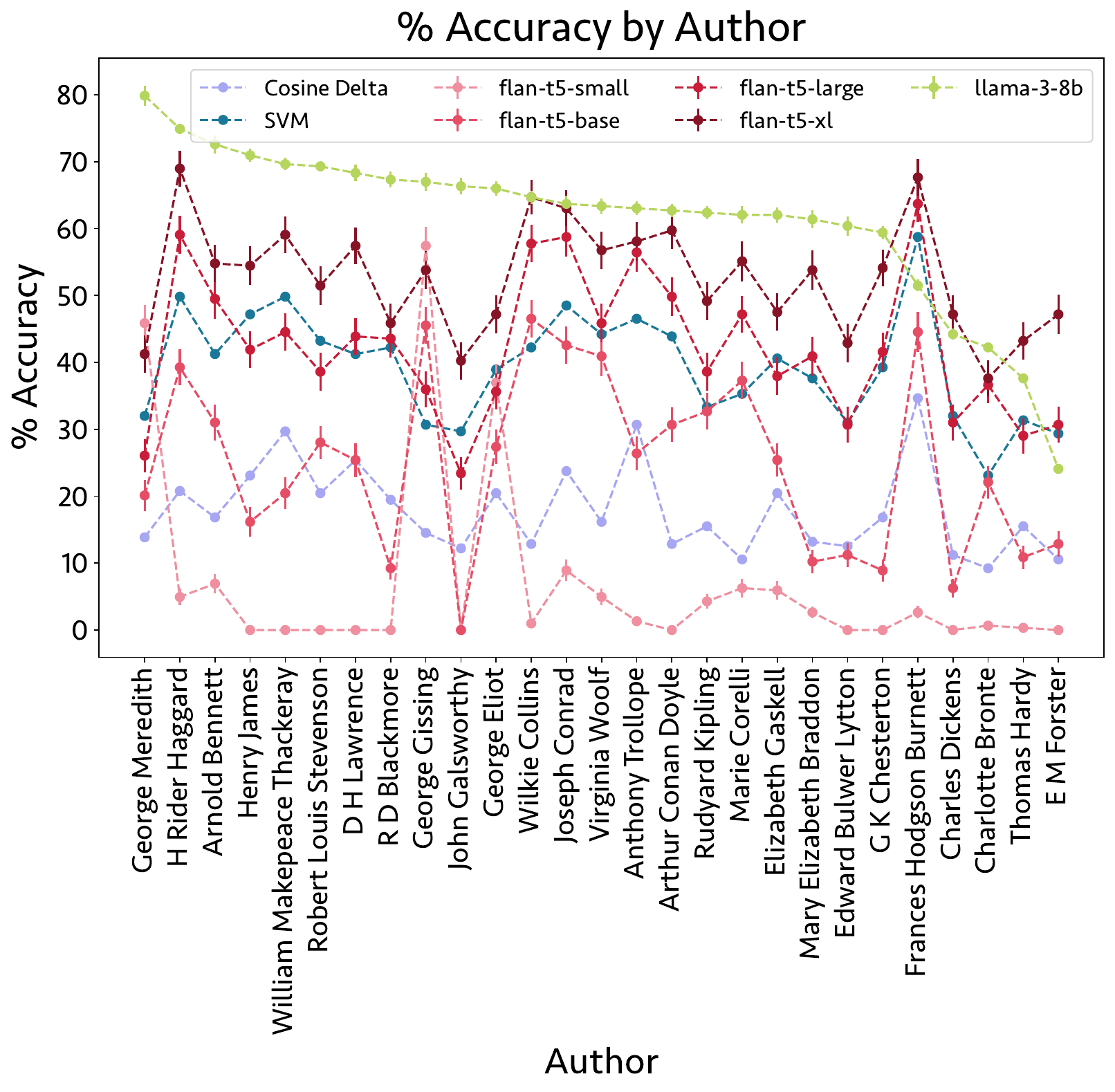} \hspace{6ex}
      \includegraphics[width=0.4\textwidth, valign=t]{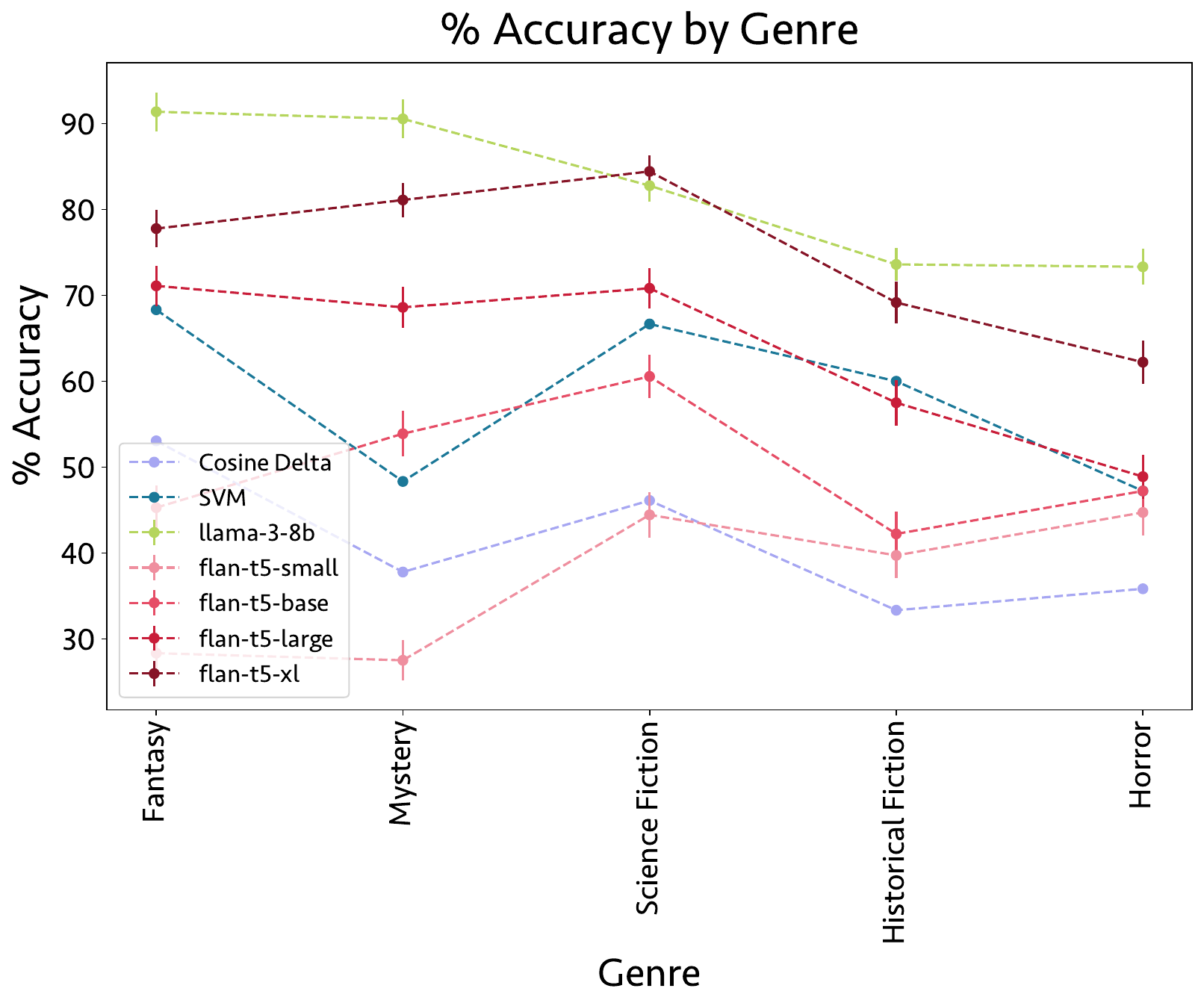}}
  \caption{Accuracy by author (left) and genre (right) for each model. Results of a single run are reported and error bars represent the standard error bootstrapped over 1,000 iterations. The x-axes are sorted by the accuracy of \texttt{llama-3-8b}.}
  \label{fig:accByClass}
\end{figure}

\subsection{Analysis by Class}

Examining results from the fine-tuned models and baselines, we see that the accuracy of each model varies more by author than by genre (Figure \ref{fig:accByClass}). Variation by class tends to lessen as the model size increases, although the difference in accuracy by author for \texttt{llama-3-8b} is greater than for \texttt{flan-t5-xl}. There is some agreement between models as to which authors are easier to classify --- for example all models but \texttt{llama-3-8b} perform comparatively very well on samples from Frances Hodgson Burnett and Joseph Conrad --- but this is not the case for all authors or all models. There are more consistent patterns across models for genre, where science fiction is the class for which each model except \texttt{llama-3-8b} performs best or second-best and horror is the least-identified class for the SVM and all transformer-based models larger than \texttt{flan-t5-base}. It is not clear why these trends exist; they may be artifacts of model performance or indicate that these classes are somehow more or less stylistically distinct than others. Overall, we find that for both genre and authorship, the difference between class-level accuracy is considerably more extreme for samples from novels withheld from training, perhaps suggesting that the model's ability to recognize author- or genre-level, and not just novel-level, signals of literary style is more variable (Appendix \ref{sec:accByClass}).

\begin{figure}[t]
  \FIG{
  \includegraphics[width=0.4\linewidth]{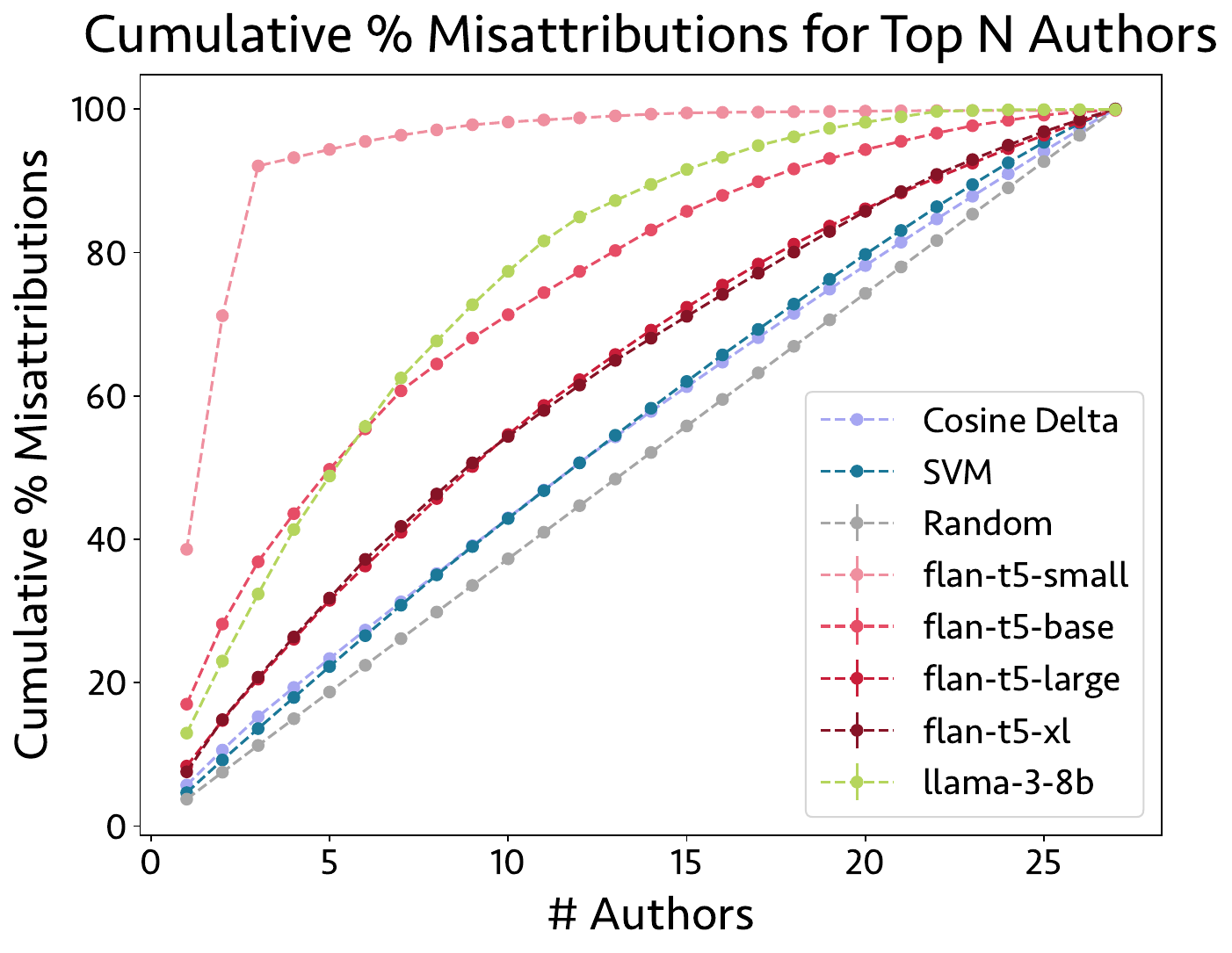}
  \hspace{6ex}
  \includegraphics[width=0.4\linewidth]{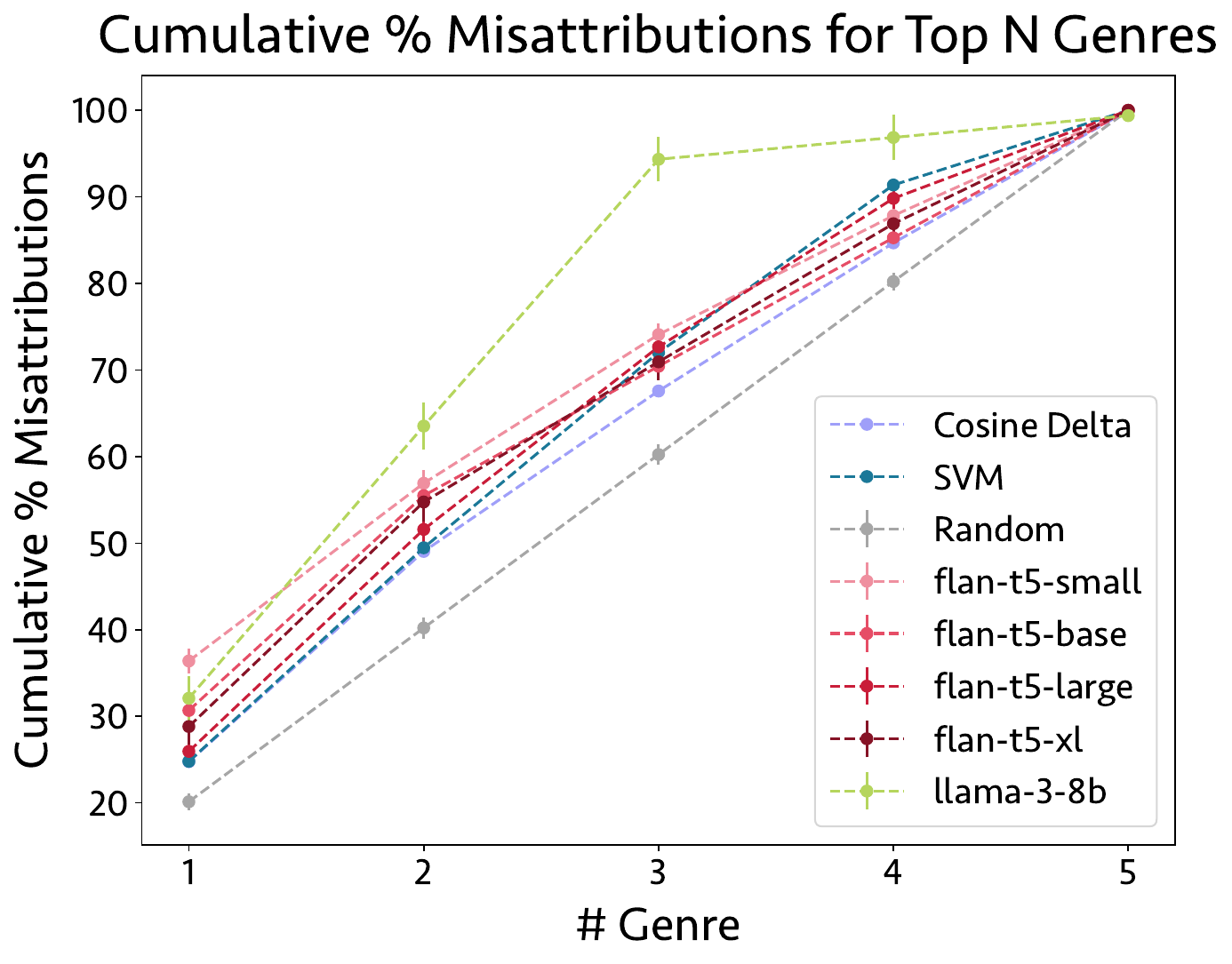}}
  {\caption{The \% of misattributions assigned to the top $n$ ``scapegoated'' authors (left) or genres (right). Results are reported for a single run and error bars represent the standard error bootstrapped over 1000 iterations.}
  \label{fig:missByClass}}
\end{figure}

Previously, \citet{hicke2023t5} found that LLMs often used particular authors as `scapegoats' when fine-tuned for authorship attribution, assigning the majority of misattributed samples to 2--3 authors. We replicate this finding for authorship attribution in novels (Figure \ref{fig:missByClass} (left)). Specifically, we find that \texttt{llama-3-8b} scapegoats more than any model except \texttt{flan-t5-small}. Although the extent to which authors are scapegoated is not consistent across models, we find that George Gissing is among the top two most scapegoated authors for each transformer-based model. \citet{hicke2023t5} found that authors with larger vocabularies and less unique language usage were more likely to be scapegoated; however, while Gissing does have the fourth lowest uniqueness score of all authors in the dataset (63.69, range: 53.02--113.88, values calculated as in \citet{hicke2023t5}), he also has the fourth-smallest vocabulary (6,468, range: 5,152--8,189). The same paper hypothesized that more famous authors may be less scapegoated, as models have more information about their writing before fine-tuning; we find that Gissing is indeed the sixth least famous author by Wikipedia length (27,013, range: 11,532--140,276). Thus, we again find evidence that less-famous authors with less-unique vocabulary usage may be more likely to be consistently scapegoated, although there are clearly other factors involved. 

In contrast to the authorship attribution results, the only model to show evidence of extreme scapegoating for genre identification is \texttt{llama-3-8b} (Figure \ref{fig:missByClass} (right)). For \texttt{llama-3-8b}, $\sim$30\% of misattributed samples are labeled as fantasy, horror, and mystery respectively, whereas only $\sim$3\% of misattributed samples are labeled as science fiction or historical fiction. There is not a large variation in genre uniqueness (range: 70.92--81.83) that explains this difference and, interestingly, the science fiction and historical fiction have the two largest vocabularies (11,369 and 10,407, range: 8,892--11,369). Perhaps the larger vocabularies of these two genres represent the usage of situation-specific terminology (e.g.\ relating to spaceships or duels) that is unlikely to appear outside of the genre, unlike a use of broader but generally applicable vocabulary by authors which may have caused the opposite trend to appear in \citet{hicke2023t5}.

\subsection{Overview}

From the classification experiments, we find that it is possible to detect multiple kinds of literary style in short text segments using LLMs. The largest, generative LLMs --- \texttt{llama-3-8b} and \texttt{flan-t5-xl} --- perform the best on both attribution tasks but variations in these models' performances reveal differences in what signals they use to recognize literary style. \texttt{llama-3-8b} appears to rely partially on memorized information, particularly about works from more popular authors, to correctly label novel excerpts, while \texttt{flan-t5-xl} demonstrates very little memorization. We also find differences between how genre- and author-level style is identified. Because the base \texttt{flan-t5-xl} model is able to produce the genre of many excerpts for which it cannot provide the author, it seems that signals of genre-level style are better known by LLMs even without memorization or fine-tuning; this makes sense, as genre is a much more generalizable concept than author. In addition, differences in the class-level accuracy of models fine-tuned for genre identification vs.\ authorship attribution and in their scapegoating suggests further differences between genre- and author-level style, or in how these styles are identified by LLMs.

\section{Ablation}

Since we have found evidence that LLMs are able to identify signals of authorial and genre-level style in short texts, in the second phase of this work we perform text ablations to probe which linguistic features models use to distinguish style. After training or fine-tuning each model on the original, unmodified texts in the training and validation datasets, we apply them to several perturbed versions of each sample: without capitalization, without punctuation, with each of nine categories of stop words masked (Appendix \ref{sec:stopwordAppendix}), with shuffled word order, with proper nouns masked, and with all modifications applied at once. We use the list of English stop words given by \texttt{nltk} and replace each stop word at the word boundary with the token \texttt{<STOP>}. Proper nouns are identified using the part-of-speech tagger from \texttt{spacy} and each proper noun is similarly replaced at the word boundary with the token \texttt{<PROPN>}. Examples of all of the perturbed versions of a single sentence are available in Appendix \ref{sec:variantAppendix}.

We then study the effect of each perturbation on the models' ability to label the samples from novels withheld from training. We focus on the samples from withheld novels as these most accurately reflect the models' ability to learn about genre-level and authorial, and not just novel-level, style.

\begin{figure}[t]
  \FIG{
  \includegraphics[width=0.4\linewidth]{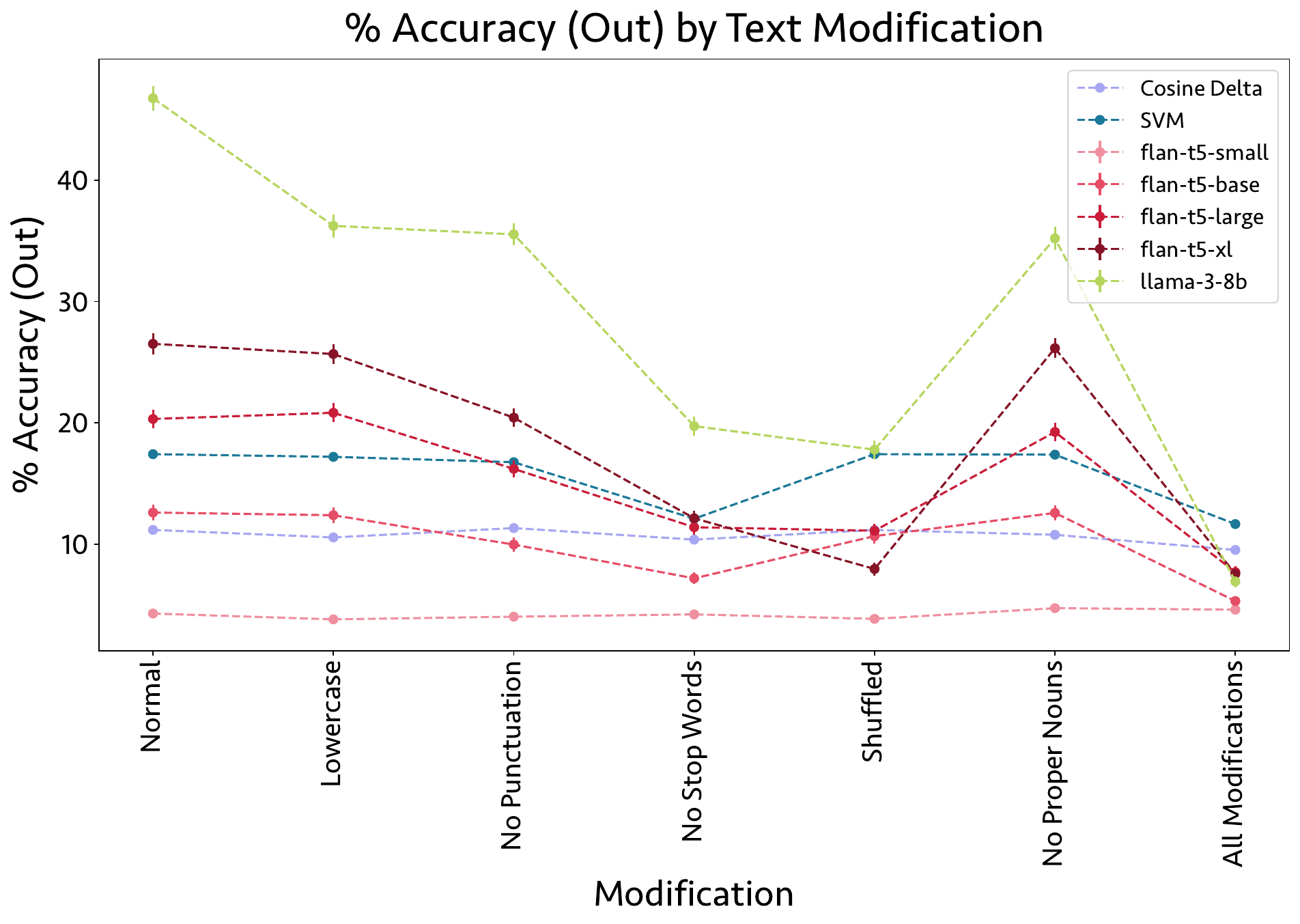}
  \hspace{6ex}
  \includegraphics[width=0.4\linewidth]{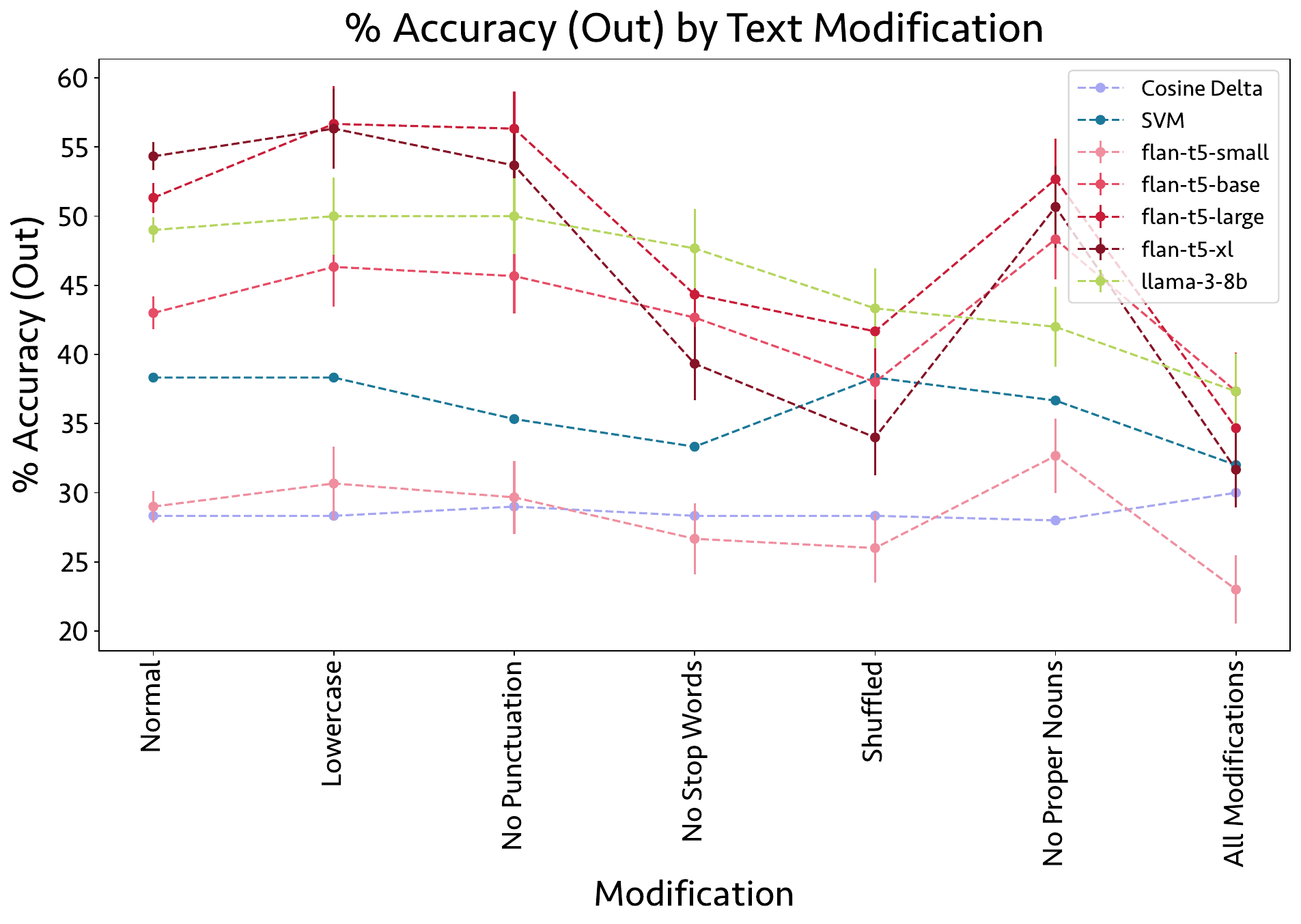}}
  {\caption{The accuracy of each model on samples from withheld novels for authorship attribution (left) and genre identification (right) across each text perturbation. Results are reported for a single run and error bars represent the standard error bootstrapped over 1000 iterations.}
  \label{fig:accByMod}}
\end{figure}

\subsection{Comparing Perturbations}

We find that the effects of each text modification become clearer in larger models and that they are relatively consistent across authorship attribution and genre identification (Figure \ref{fig:accByMod}). However, the accuracy drops slightly when capitalization and punctuation are removed in the Flan-T5 and Llama models for authorship attribution, but these disruptions barely affect or improve these models' performance for genre identification. These results indicate that capitalization and punctuation patterns may carry information about authorship, but do not indicate genre; this makes intuitive sense as authors seem more likely develop specific punctuation of capitalization habits than entire genres. The performance of models degrades somewhat when proper nouns are removed for both tasks, but this effect is relatively small for samples from withheld novels (on average 9.28\% for \texttt{llama-3-8b} and 4.02\% for \texttt{flan-t5-xl}). 

For all of the large language models, the most impactful modifications are masking stop words and shuffling word order. In fact, when the input texts are shuffled the performance of \texttt{flan-t5-xl} drops below that of the SVM, a bag of words model, for both stylistic identification tasks. This demonstrates how heavily reliant the models, particularly larger models, are on sequence-level information and how strongly their representations of style rely on word order.

The apparent impact of stop words initially appears consistent with the findings of previous stylometric studies tracing back to \citet{mosteller1972deciding}, which argue that common function words play a large role in defining authorial style. However, we further evaluate the two highest performing models on an alternate perturbed dataset where, instead of masking stop words, we mask a number of random words in each sample equal to the number of stop words in that sample. Performance on this dataset is lower for both tasks and both models (by $\sim$8\% for \texttt{llama-3-8b} and $\sim$0.5\% for \texttt{flan-t5-xl} on withheld data) than when only stop words are masked. Therefore, while stop words clearly contribute to distinguishing authorial and genre-level style, as past research has suggested, these results suggest that LLMs make equal or greater use of less-frequent words.

\begin{figure}[t]
  \FIG{
  \includegraphics[width=0.8\linewidth]{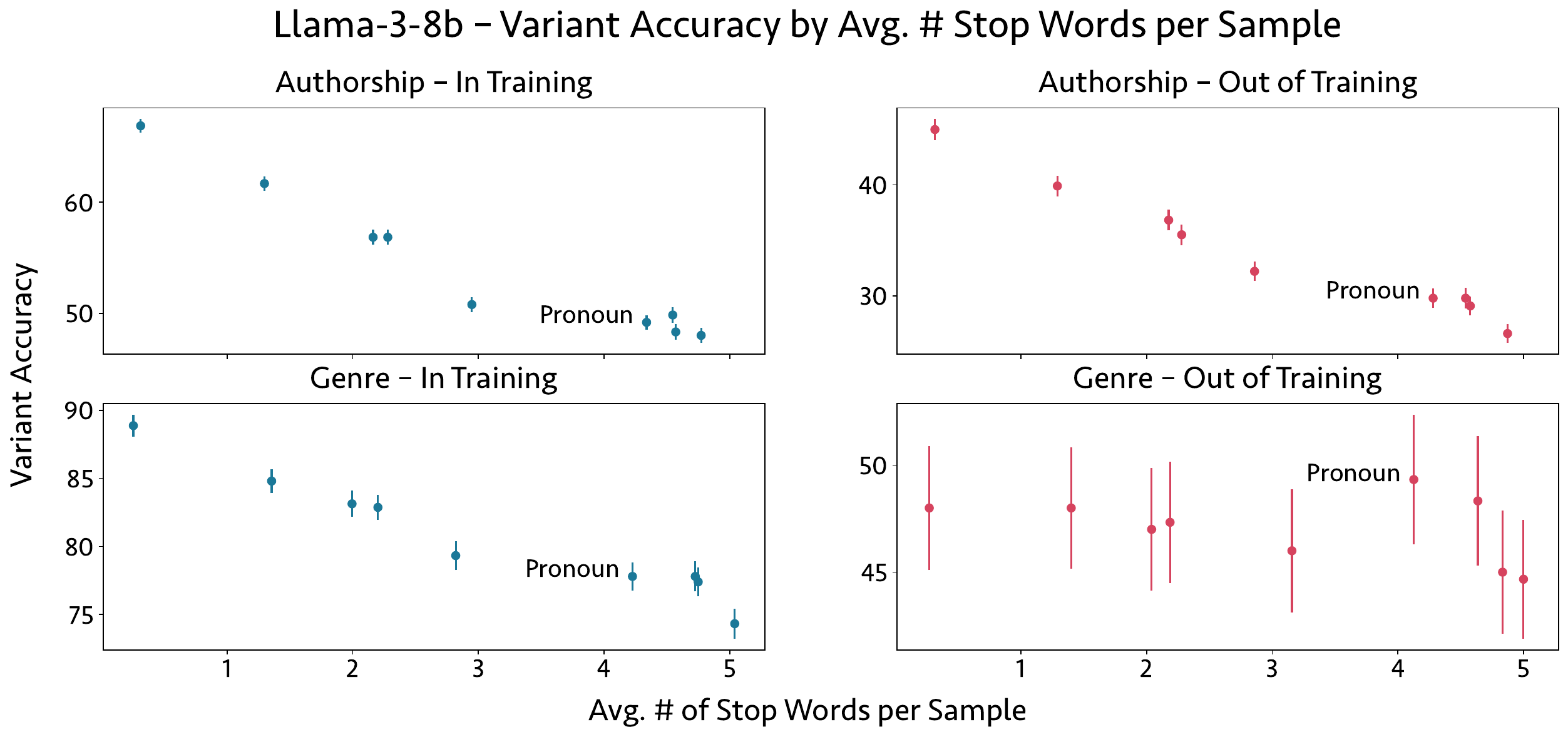}
  \includegraphics[width=0.8\linewidth]{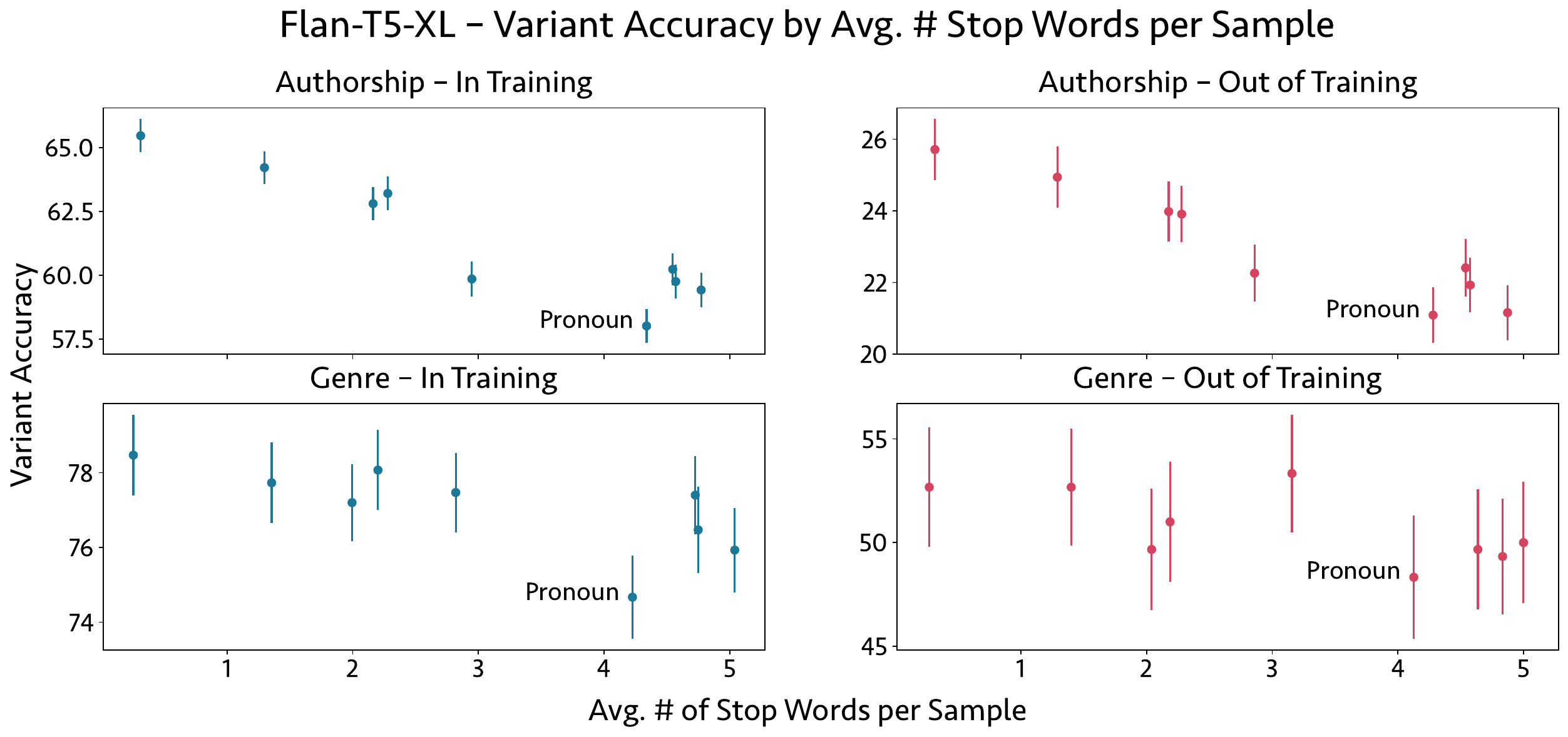}}
  {\caption{\texttt{llama-3-8b} (top) and \texttt{flan-t5-xl}'s (bottom) accuracy when stop words of a certain part of speech are masked by average number of those stop words per sample for both tasks. Results are reported for a single run and error bars represent the standard error bootstrapped over 1000 iterations. The pronoun data are labeled to provide context for in-text analysis.}
  \label{fig:stopWordsAcc}}
\end{figure}

Because masking stop words for some parts-of-speech affects more words than others, we next look at the relationship between the average number of masked words per sample and model accuracy for each stop word variant. For authorship attribution in \texttt{llama-3-8b} and \texttt{flan-t5-xl} there is a significant negative relationship (all Pearson's R < -0.9, $p < 5\times10^{-4}$) between the number of words masked and model accuracy for both out-of-training and in-training novel samples (Figure \ref{fig:stopWordsAcc}).
This indicates that the differences in model accuracy between each stop word variant are largely due to the number of stop words being masked and not the importance of certain types of stop words for authorship attribution.

However, for genre identification in both models there is only a significant relationship between the number of words masked and variant accuracy for novels in training (Pearson's R < -0.7, $p < 0.03$). Therefore, we hypothesize that some stop word variants do have a disproportionate effect for models when generalizing to genre-level style; in particular, we find that pronouns have an unexpectedly large effect for \texttt{flan-t5-xl} whereas conjunctions have an unexpectedly large impact and pronouns and prepositions have an unexpectedly small effect on \texttt{llama-3-8b}.

\begin{figure}[t]
  \FIG{
  \includegraphics[width=0.8\linewidth]{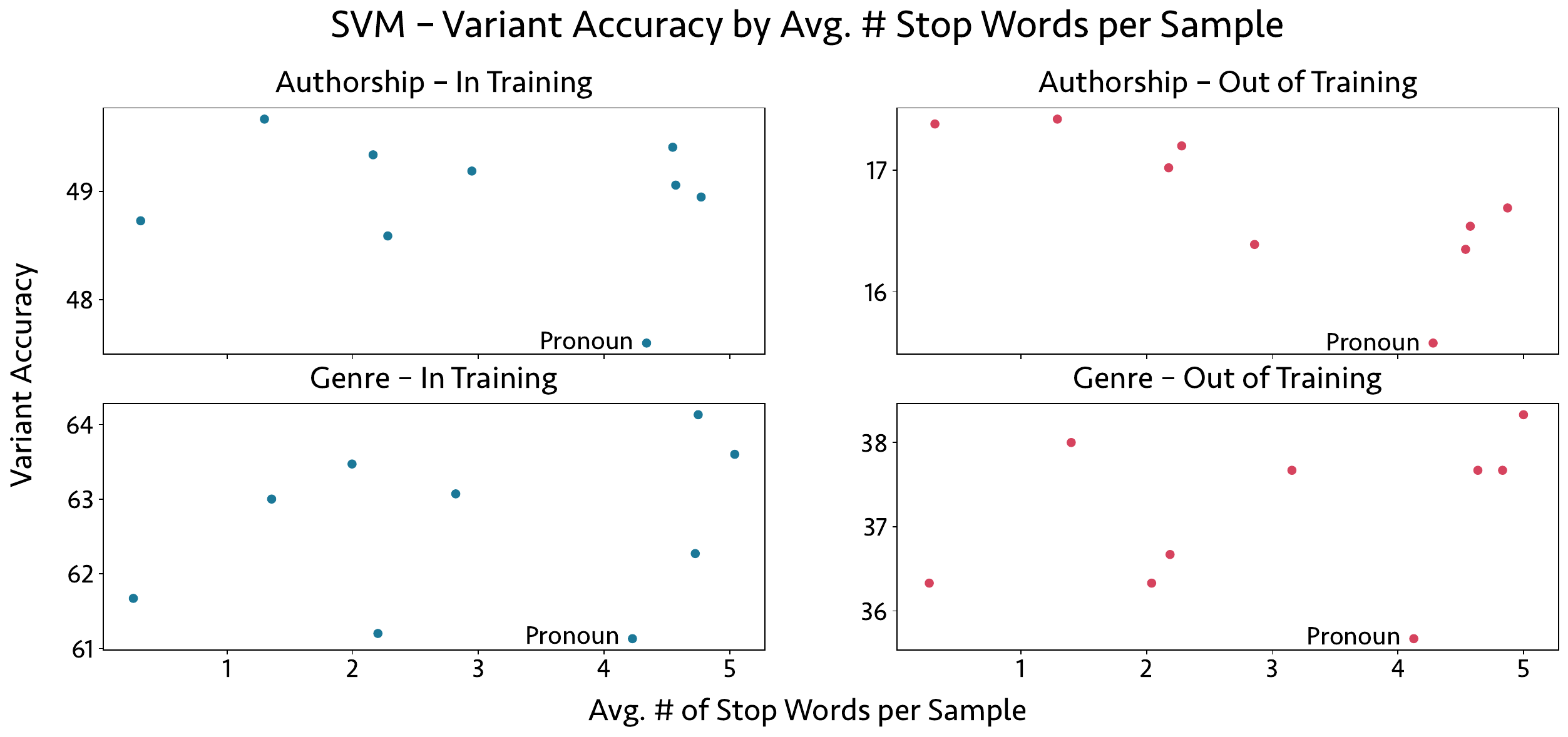}
  \includegraphics[width=0.8\linewidth]{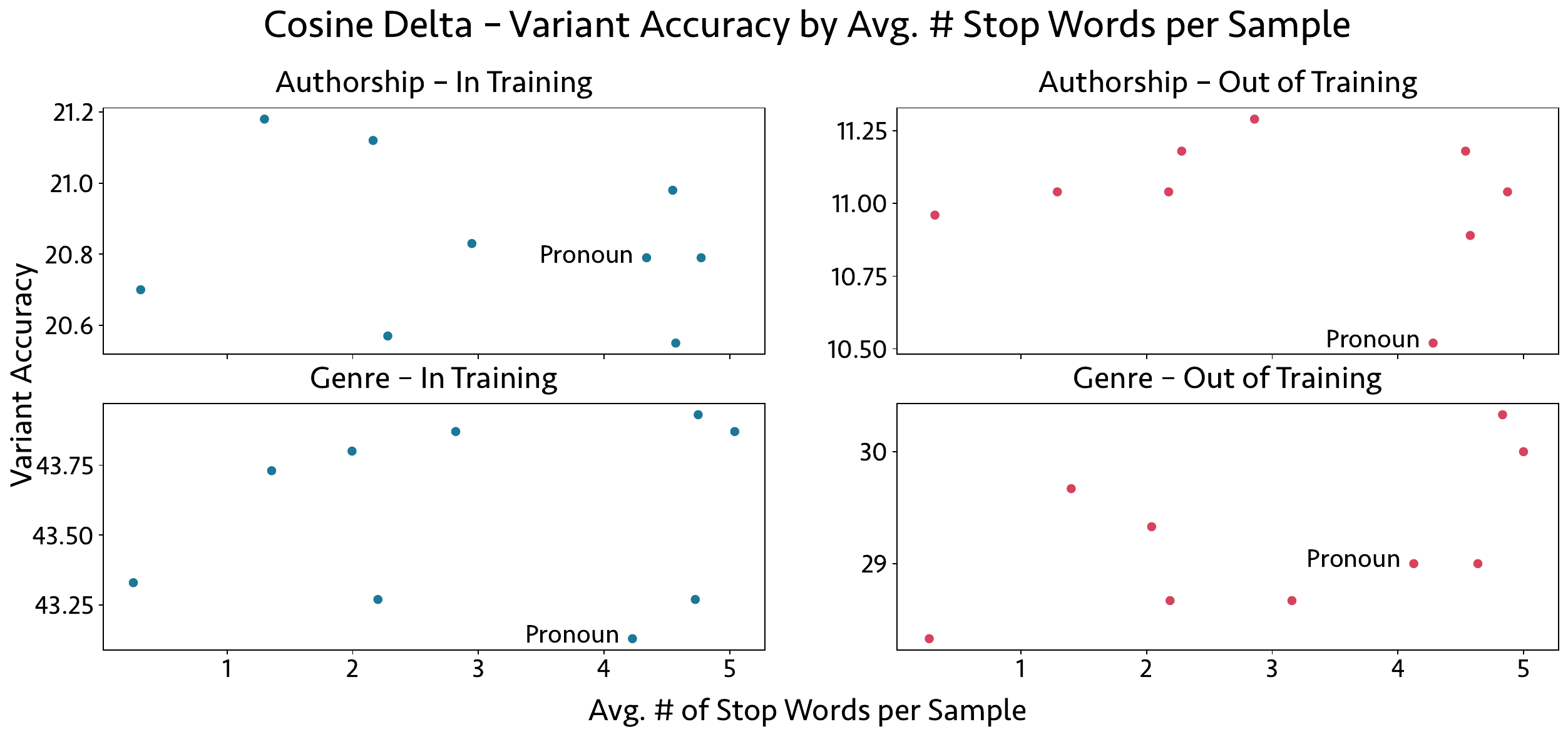}}
  {\caption{The SVM (top) and cosine delta's (bottom) accuracy when stop words of a certain part of speech are masked by average number of those stop words per sample for both tasks. The pronoun data are labeled to provide context for in-text analysis.}
  \label{fig:stopWordsAccBase}}
\end{figure}

Overall, regardless of correlation with the number of words masked, we find that pronouns have the largest effect on \texttt{flan-t5-xl} on both tasks for both samples from novels in-training and out-of-training (Figure \ref{fig:stopWordsAcc}). Pronouns also have a large impact on the bag-of-words baselines; they have the largest effect on both tasks for the SVM and on authorship attribution for cosine delta (Figure \ref{fig:stopWordsAccBase}). In contrast, for \texttt{llama-3-8b} pronouns do not have a disproportionate effect on accuracy; they are no more or less impactful than any other stop word category conditioned on number of words masked. However, this may be because \texttt{llama-3-8b} is relying on memorized information and therefore masking stop words primarily impacts the model's ability to match inputs with memorized texts. Because pronouns disproportionately affect the models which have not shown signs of memorized information, we hypothesize that they are important for detecting literary, particularly authorial, style.

\subsection{Overview}

These ablation experiments allow us to infer information on the relative importance of several syntactic and semantic text properties. We find that some stylistic choices, like capitalization and punctuation, affect the more fine-grained authorial style but have less impact on the more generalized genre-level style. Proper nouns have little impact overall on models' abilities to identify texts from out-of-training novels, likely because they are usually novel-specific. On the other hand, word order proves to be very important in identifying both forms of literary style, especially for the \texttt{flan-t5-xl}. Finally, we find evidence that stop words are not as important for distinguishing literary style as previous work has suggested; despite this, for models that do not show evidence of text memorization pronouns appear to be impactful in characterizing literary, particularly authorial, style.

\section{Probing}

In the third phase of this study, we probe the fine-tuned \texttt{flan-t5-xl} model for evidence of which features contribute to representations of literary style. We do not provide similar probes of \texttt{llama-3-8b} because the quantized version of the model we use does not expose the intermediate values necessary for these analyses.

\subsection{Training Influence}

Our first internal probe attempts to discern which input features, here words, most influence the style-detection models. We hypothesize that the words with the greatest impact on model behavior will carry the most information about style. To measure the impact of words, we use Shapley values \citep{NIPS2017_8a20a862}, a metric from game theory that measures the contribution of each input feature to an outcome. Specifically, for this probe we use a gradient-based implementation for calculating Shapley values available through the \texttt{inseq} package \citep{sarti-etal-2023-inseq} to examine the impact of each word in the input on each token in the output representing an author name or genre. We strip punctuation and capitalization from features during analysis and remove tokens from the input prefixes (\texttt{GENRE: <extra\_id\_0>} and \texttt{AUTHOR: <extra\_id\_0>}). We then look at the most impactful words by mean and summed Shapley value over all excerpts in the test datasets.

Unfortunately, the resulting Shapley values do not provide clearly interpretable information about style. The most prominent words by mean Shapley value appear very rarely in the input texts and consist primarily of proper nouns and other uncommon words (Appendix \ref{sec:shapValues}). While this informs us that rare or uncommon words can be highly impactful in attributing the individual samples containing them, it does not allow us to draw broader conclusions about what features \texttt{flan-t5-xl} makes use of. In contrast, the most impactful words by summed Shapley value appear to be the most frequent words in the corpus. Many pronouns appear among the most impactful words by summed Shapley value for both genre indentification and authorship attribution --- more than any other stop word --- but this may simply reflect their frequent usage. Overall, the Shapley values demonstrate that words used at a variety of frequencies can be impactful for style identification, but do not provide further insight.

\subsection{Cross Attention}

Our second probe examines the cross attention paid by each token representing an author name or genre (from here on \textit{answer tokens}) in an output to every token in the input sample. Here, we again hypothesize that the most influential words in the input will hold the most information about style, and use attention as a proxy for influence. For every sample, we create a version of the output for each correct and incorrect class (for example, a sample from a horror novel will have output versions created reporting that it is from each of the five genres) and then extract the attention paid by the answer tokens to the input. We analyze the tokens with the largest summed and mean attention values across all correct and incorrect answers. 

As with Shapley values, the highest mean attention is paid to rare or unusual words, while the most frequently used words rank highly by summed attention (Appendix \ref{sec:authorAttentionOverall} and \ref{sec:genreAttentionOverall}). In order to extract more interpretable results from these values and compare cross-attention values, we use a modified version of the \textit{fightin' words} algorithm \citep{monroe2008fightin},\footnote{\url{https://github.com/jmhessel/FightingWords}} a Bayesian text analysis method that describes which words distinguish two corpora. To adapt the algorithm for use with attention values, we simply replace the standard matrix of word counts with a matrix of summed attention values. Because samples with incorrect answer tokens are evaluated 26 times (authorship attribution) or 4 times (genre identification) more frequently with than those with correct answer tokens,  we standardize the summed attention values from these samples.

We first use the modified fightin' words algorithm to compare the summed attention values for each class to the values for all other classes in that dataset (e.g.\ Charles Dickens vs.\ all other authors). Then we re-rank the input tokens by their fightin' words scores to find the twenty most and least distinctive words for each genre or author (Appendix \ref{sec:authorAttentionAuthors} and \ref{sec:genreAttentionGenre}). Next, we count which token types feature frequently in these lists, counting a token each time it appears and allowing for double counting. From this analysis, we find that punctuation makes up 16\% of the most distinctive tokens for authors and 12\% for genre. Again, these stylistic elements appear to be slightly more important for distinguishing authorial style than genre-level style. Stop words make up 26.5\% and 23.7\% of the most distinctive tokens for authors and genres respectively. For both genre identification and authorship attribution, pronouns appear the most of any stop word type in these lists (11.5\% and 12.96\%), once more reinforcing their importance. Proper nouns also appear frequently in the distinctive words for authors and occasionally for genres; however, these results may be affected by results from excerpts from in-training novels, where proper nouns may be significant aids in attribution.

The pronouns in the distinctive word lists for each author appear to most frequently distinguish between commonly used points of view (e.g.\ first person vs.\ third person). However, they also seem to provide information about the subjects and characters the authors frequently discuss. We use the fightin' words algorithm to run a further comparison of the attention paid to input tokens by answer tokens representing correct female and male authors. The two tokens that most strongly characterize female authors' language by this metric are \textit{\_her} and \textit{\_she}; the token \textit{\_She} is the only other pronoun to appear in the list of fifty most distinctive words for female authors. In contrast, the only pronouns to appear in the set of fifty most distinctive tokens for male authors are \textit{\_we}, \textit{\_us}, \textit{\_our}, and \textit{\_his}. These results demonstrate that female pronouns are more stylistically important for attributing texts from female authors, perhaps suggesting that female authors more frequently discuss women or do so distinctly from male authors. Interestingly, the results also show that first-person plural pronouns are similarly stylistically significant for attributing texts from male authors.

Next, we use the fightin' words algorithm to compare the attention paid to input tokens when the correct and incorrect classes are represented by the answer tokens. For the authorship attribution task, we see that proper nouns and common subword tokens appear frequently in the fifty most distinctive tokens for correct authors (Appendix \ref{sec:authorAttentionCorrect}). The only stop word present in this list is \textit{\_the}. In contrast, for genre identification, the fifty tokens most distinctive of correct answers are mostly punctuation, stop words, and simple subword tokens (Appendix \ref{sec:genreAttentionCorr}). \textit{\_the} is the most distinctive word for correct genres. 
While the word \textit{\_the} is unlikely to ground compelling theories of genre by itself, it may be standing in implicitly for features like the use of concrete nouns.

The tokens most associated with incorrect answers for both authorship attribution and genre identification contain many capitalized words, which likely represent the the first words in samples. Many of these are stop words. This could indicate that the fine-tuned \texttt{flan-t5-xl} models pool attention in the first word of a sentence when confused or uncertain; this hypothesis is supported by the results of \citet{xiao2023streamingllm}, which  show that the first words of sequences act as attention sinks in decoder-only models during dialogue exchanges.

The results of the cross-attention analysis again suggest that there are differences in authorial and genre-level style. Proper nouns and specific topical words appear to be more impactful in forming authorial style, whereas more common words and small subword tokens are comparatively more important for characterizing genre-level style. Perhaps this is because authors use a relatively set vocabulary in their works, but the same is not true for a genre made up of works by many authors with varied vocabularies. Thus, in distinguishing genre, the presence of very specific terms may not be a reliable signal, whereas the frequency with which particular sets of more common words are used plays a larger role.

\subsection{Contextual Embeddings}

The previous probes have treated words as unigrams. In our third probe, we explore whether \texttt{flan-t5-xl}'s understanding of style is contextual; specifically, we examine whether \textit{how} words are used matters as well as which words are used. We hypothesize that both vocabulary and word usage will contribute to authorial and genre-level style. 

To test this hypothesis, we model the language of each class (author or genre) in our datasets by creating average embeddings representing how words are used by that class. Specifically, we pass every excerpt from the training datasets through the base \texttt{flan-t5-xl} model and create a contextual embedding of each word in that excerpt by averaging the embeddings of the word's component tokens from the last four layers of the model's encoder stack. We then create a class-level embedding for each word by averaging the embeddings of each instance of the word from that class in the training dataset. This results in a dictionary for each class containing an average embedding representing for each word in the training dataset vocabulary describing how that word is used by that class. We create a representation of the `average' language usage in the training dataset by averaging across the class-level representations of a word for every class that used that word.

To compare how unique each class's language usage is, we find the average cosine similarity between the class's usage of each word and the training dataset's average usage of that word, adding zero when a word in the training dataset vocabulary was never used by the class. Similarly, we compare the language usage in a class's test data to its training data by finding the average cosine similarity between each word's embedding in a test sample and the embedding of that word created from the class's training data and summing this average cosine similarity for each excerpt in the class's test data.

We find that, for authorship attribution, the fine-tuned \texttt{flan-t5-xl} model performs better the more unique an author's language is, compared to the training dataset norm, and the more similar their samples in the test dataset are to their samples in training. Specifically, we find a significant negative relationship between author-level accuracy and the similarity of an author's word usage in training to the average training dataset language (Pearson's R: -0.57, $p < 5\times10{-3}$) and a significant positive relationship between author-level accuracy and the similarity of the author's test and training word embeddings (Pearson's R: 0.66, $p < 5\times10^{-4}$). These correlations suggest that contextual word usage does help \texttt{flan-t5-xl} to distinguish between authorial styles, with more distinct authors by this metric becoming easier to identify. 

However, there are no significant correlations between genre-level accuracy and similarity between average training embeddings and genre-specific training embeddings ($p > 0.6$) or similarity between genre-specific training and test embeddings ($p > 0.8$). This indicates that contextual word usage may do less to differentiate genres than authors; perhaps, this is again because genres are contributed to by many authors who may all have distinct word usage patterns.

\subsection{Overview}

From our probing experiments, we find that both frequently and infrequently used words can be useful for distinguishing between literary styles. Whereas uncommonly used words may be more individually powerful in determining authorial or genre-level style, common words that carry small stylistic signals can become significant over a corpus. We also find further evidence of differences between genre-level and authorial style. The results of the cross-attention and contextual embedding probes suggest that more uncommon words and words' contextual usages are impactful in characterizing authorial style, but not genre-level style. These features may be less impactful for genre-level style because genres are made up of works by a large variety of authors who can all use different vocabularies in different ways. Finally, we again find evidence that pronouns help to distinguish literary styles, both between authors and, notably, between male and female authors.

\section{Conclusion}

We find that generative large language models are able to recognize signs of authorial and genre-level style in very short segments of text. This proves both that stylistic signals exist at the sentence level and that models are capable of recognizing such subtle linguistic patterns that are largely invisible to the human reader. However, differences exist between the information different model families use to classify texts; whereas Flan-T5 appears to learn stylistic representations throughout fine-tuning, Llama-3 leverages some memorized information from pre-training. 

We also find differences between the signal for authorial and genre-level style. 
Unsurprisingly, the small, grammatical, mostly unconscious word choices typical of author signatures, such as punctuation or details of conceptual word usage, are less impactful for genre.
Nevertheless, we do find that word order and some high-frequency words, specifically pronouns, are significant for characterizing both authorial and genre-level style.

This work is a step towards a data-driven but human-interpretable analysis of the characteristics that define author and genre style.
Language models can help bridge the gap between qualitative ``I know it when I see it'' distinctions and minimal, overly mechanistic stylometric analyses.
In future work, we hope to develop more sophisticated methods for probing for and characterizing human-recognizable elements of style beyond individual words.
We will also benefit from expanding our genre corpus to a larger number of works, and, if possible, extending to more recent 20th and 21st century works following the growth of commercial genre fiction.

\begin{Backmatter}

\paragraph{Acknowledgments}
We would like to thank Axel Bax, Federica Bologna, Sil Hamilton, Kiara, Liu, Noam Ringach, Rosamond Thalken, Andrea Wang, Matthew Wilkens, and Shengqi Zhu for their thoughtful feedback. This work was supported by the NEH project AI for Humanists.

\paragraph{Data availability statement}

The data used in this project can be found at the following anonymized Github repository: \url{https://anonymous.4open.science/r/looking-for-the-inner-music-93FD/README.md}+. This material, in conjunction with the descriptions of methods provided in this paper, allow for replication of the reported results.

\paragraph{Ethical standard}
The authors affirm this research did not involve human participants.

\paragraph{Author~contribution}
Rebecca M.\ M.\ Hicke: Conceptualization, Data curation, Formal analysis, Investigation, Methodology, Software, Visualization, Writing -- original draft, Writing -- review \& editing. David Mimno: Conceptualization, Funding acquisition, Methodology, Resources, Supervision, Writing -- original draft, Writing -- review \& editing. All authors approved the final submitted draft.

\paragraph{Funding statement}

This project was funded by the NEH project AI for Humanists.

\paragraph{Competing interests}

None.

\bibliographystyle{chicago}
\bibliography{custom.bib}





\appendix

\section{Corpora Contents}
\label{sec:corporaAppendix}

Novels withheld from the training data are italicized.

\subsection{Authorship Attribution Corpus}
\label{sec:authorCorpusAppendix}

\textbf{Arnold Bennett:} The Grand Babylon Hotel (1902), \emph{Helen with the High Hand} (1910), Imperial Palace (1930)\\
\newline
\textbf{R.\ D.\ Blackmore:} Lorna Doone (1869), \emph{Erema} (1877), Springhaven (1887)\\
\newline
\textbf{Mary Elizabeth Braddon:} \emph{Lady Audley's Secret} (1862), Fenton's Quest (1871), Phantom Fortune (1883)\\
\newline
\textbf{Charlotte Bronte:} Jane Eyre (1847), Shirley (1849), \emph{Villette} (1853)\\
\newline
\textbf{Frances Hodgson Burnett:} \emph{Little Lord Fauntleroy} (1886), A Little Princess (1905), The Secret Garden (1911)\\
\newline
\textbf{G.\ K.\ Chesterton:} The Napoleon of Notting Hill (1904), \emph{The Man Who Was Thursday} (1908), The Innocence of Father Brown (1911)\\
\newline
\textbf{Wilkie Collins:} Basil (1852), The Woman in White (1860), \emph{The Legacy of Cain} (1889)\\
\newline
\textbf{Joseph Conrad:} Almayer's Folly (1895), Nostromo (1904), \emph{The Rover} (1923)\\
\newline
\textbf{Marie Corelli:} A Romance of Two Worlds (1886), The Sorrows of Satan (1895), \emph{Innocent} (1914)\\
\newline
\textbf{Charles Dickens:} \emph{Oliver Twist} (1839), Bleak House (1853), Great Expectations (1861)\\
\newline
\textbf{Arthur Conan Doyle:} Micah Clarke (1889), The Hound of the Baskervilles (1902), \emph{The Lost World} (1912)\\
\newline
\textbf{George Eliot:} Adam Bede (1859), Felix Holt, The Radical (1866), \emph{Daniel Deronda} (1876)\\
\newline
\textbf{E.\ M.\ Forster:} \emph{Where Angels Fear to Read} (1905), A Room with a View (1908), Howards End (1910)\\
\newline
\textbf{John Galsworthy:} \emph{The Man of Property} (1906), Saints Progress (1919), Over the River (1933)\\
\newline
\textbf{Elizabeth Gaskell:} Ruth (1855), Sylvia's Lovers (1863), \emph{Wives and Daughters} (1865)\\
\newline
\textbf{George Gissing:} The Unclassed (1884), The Odd Women (1893), \emph{Will Warburton} (1903)\\
\newline
\textbf{H.\ Rider Haggard:} King Solomon's Mines (1885), \emph{She: A History of Adventure} (1887), She and Allan (1921)\\
\newline
\textbf{Thomas Hardy:} Far from the Madding Crowd (1874), \emph{Tess of the D'Ubervilles} (1891), Jude the Obscure (1895)\\
\newline
\textbf{Henry James:} Roderick Hudson (1875), \emph{The Tragic Muse} (1890), The Ambassadors (1903)\\
\newline
\textbf{Rudyard Kipling:} The Light that Failed (1891), \emph{Captains Courageous} (1896), Kim (1901)\\
\newline
\textbf{D.\ H.\ Lawrence:} The White Peacock (1911), Women in Love (1920), \emph{The Plumed Serpent} (1926)\\
\newline
\textbf{Edward Bulwer Lytton:} My Novel (1853), What Will He Do With It (1858), \emph{Kenelm Chillingly} (1873)\\
\newline
\textbf{George Meredith:} The Ordeal of Richard Feverel (1859), The Adventures of Harry Richmond (1871), \emph{The Amazing Marriage} (1895)\\
\newline
\textbf{Robert Louis Stevenson:} \emph{Treasure Island} (1883), The Black Arrow (1888), Catriona (1893)\\
\newline
\textbf{William Makepeace Thackeray:} The History of Pendennis (1850), The History of Henry Esmond (1852), \emph{The Virginians} (1859)\\
\newline
\textbf{Anthony Trollope:} \emph{The Warden} (1855), Phineas Finn (1869), Ayala's Angel (1881) \\
\newline
\textbf{Virginia Woolf:} Night and Day (1919), \emph{To the Lighthouse} (1927), The Years (1937)

\subsection{Genre Identification Corpus}
\label{sec:genreCorpusAppendix}

\textbf{Fantasy:} Gulliver's Travels by Jonathan Swift (1726), Alice's Adventures in Wonderland by Lewis Carroll (1871), \emph{She: A History of Adventure by H.\ Rider Haggard} (1887), Lilith: A Romance by George MacDonald (1895), Phantastes by Ingersoll Lockwood (1893), The King of Elfland's Daughter by Lord Dusany (1924) \\
\newline
\textbf{Historical Fiction:} Twenty Years After by Alexandre Dumas and August Maquet (1845), A Tale of Two Cities by Charles Dickens (1859), Les Misérables by Victory Hugo, translated by Isabel Florence Hapgood (1862), War and Peace by Leo Tolstoy, translated by Aylmer Maude and Louise Maude (1868), Middlemarch by George Eliot (1871), \emph{The Prince and the Pauper by Mark Twain} (1881) \\
\newline
\textbf{Horror:} Varney the Vampire by James Malcolm Rymer (1845), \emph{Carmilla by J.\ Sheridan Le Fanu} (1872), The Strange Case of Dr.\ Jekyll and Mr.\ Hyde by Robert Louis Stevenson (1886), Dracula by Bram Stoker (1897), The Turn of the Screw by Henry James (1898), The House of the Vampire by George Sylvester (1907) \\
\newline
\textbf{Mystery:} A Study in Scarlet by Arthur Conan Doyle (1887), The Mystery of the Yellow Room by Gaston Leroux (1907), Whose Body? by Dorothy L.\ Sayers (1923), The Murder of Roger Ackroyd by Agatha Christie (1926), \emph{Mystery at Lynden Sands by J.\ J.\ Connington} (1928), The Woman in Black by Susan Hill (1947) \\
\newline
\textbf{Science Fiction:} Twenty Thousand Leagues Under the Sea by Jules Verne (1869), Flatland: A Romance of Many Dimensions by Edwin A.\ Abbot (1884), The War of the Worlds by H.\ G.\ Wells (1898), \emph{The Land that Time Forgot by Edgar Rice Burroughs} (1918), Triplaneatry by E.\ E.\ ``Doc'' Smith (1934), Down and Out in the Magic Kingdom by Cory Doctorow (2003)

\section{Model and Package Information}
\label{sec:parameterAppendix}

\subsection{Licensing}

The pre-trained Flan-T5 models and Unsloth AI's Llama-3 8b model are available under the Apache 2.0 license. Jack Hessel's Fightin' Words implementation is avaible under an MIT license. All Python packages used are available with the MIT, BSD-3-Clause, or Apache 2.0 licenses except for \texttt{numpy}, \texttt{matplotlib}, and \texttt{cuda} which are licensed ``AS IS.'' All texts used in the experiments are available under public domain in the United States as of 2024.

\subsection{Package Version and Settings}

Each package is used with default settings. The most updated version of each package as of June 1, 2024 is used for these experiments.

\subsection{GPU Hours}

All of the large language models were trained on GPU. The majority of evaluation occurred on CPUs, but \texttt{llama-8-3b} and \texttt{flan-t5-xl} were evaluated on GPU for speed. The word embedding analysis and Shapley values were also produced on GPU. Overall, approximately 2 weeks of GPU hours on an Nvidia A6000 were used and an equivalent number of CPU hours.

\subsection{Model Sizes}

\begin{table}[hbt!]
\TBL{\label{tab:modelSizes}}{\begin{fntable}\tabcolsep=2pt
\begin{tabular}{|c|c|}
\toprule
\textbf{Model} & \textbf{\# Parameters} \\
\midrule
\texttt{flan-t5-small} & 60M \\
\texttt{flan-t5-base} & 220M \\
\texttt{flan-t5-large} & 770M \\
\texttt{flan-t5-xl} & 3B \\
\texttt{llama-3-8b} & 8b \\
\botrule
\end{tabular}
\end{fntable}}
\end{table}

\subsection{Flan-T5 Fine-tuning}
\label{sec:flanParameterAppendix}

\textbf{Training Parameters:}
The evaluation strategy is ``epoch.'' The learning rate is 2e-5 and the weight decay is 0.01. The per device train and eval batch sizes are 32 for the \texttt{small}, \texttt{base}, and \texttt{large} models and 16 for the \texttt{xl} model. The save total limit is 3 and the number of train epochs is 10. Gradient checkpointing is set to \texttt{True} and the maximum sequence length is 120.

\subsection{Llama-3 Fine-tuning}
\label{sec:llamaParameterAppendix}

\textbf{Training Parameters:}
Packing is set to \texttt{False}. The learning rate is 2e-5 and the weight decay is 0.01/ The per device train and eval batch sizes are 24. The number of train epochs is 10. The optimizer is Adam-W 8bit and the learning rate scheduler is linear. The maximum sequence length is 120. fp16 is set to \texttt{False} and bf16 is set to \texttt{True}. \\
\: \\
\noindent
\textbf{LoRA Parameters:}
The target modules are \texttt{q\_proj}, \texttt{k\_proj}, \texttt{v\_proj}, \texttt{o\_proj}, \texttt{gate\_proj}, \texttt{up\_proj}, and \texttt{down\_proj}. The LoRA parameters $r$ and $\alpha$ are both set to 16. Dropout is 9 and the bias is \texttt{None}.

\section{Example Variants}
\label{sec:variantAppendix}

\textbf{Normal:} ``Then come with me,'' said Mrs. Sowerberry: taking up a dim and dirty lamp, and leading the way upstairs; ``your bed's under the counter.''\\
\newline
\textbf{Lowercase:} ``then come with me,'' said mrs. sowerberry: taking up a dim and dirty lamp, and leading the way upstairs; ``your bed's under the counter.''\\
\newline
\textbf{No Punctuation:} Then come with me said Mrs Sowerberry taking up a dim and dirty lamp and leading the way upstairs your beds under the counter\\
\newline
\textbf{No Stop Words:} ``<STOP> come <STOP> <STOP>,'' said Mrs. Sowerberry: taking <STOP> <STOP> dim <STOP> dirty lamp, <STOP> leading <STOP> way upstairs; ``<STOP> bed'<STOP> <STOP> <STOP> counter.''\\
\newline
\textbf{Shuffled:} way said up dirty the dim Sowerberry: bed's leading taking and lamp, upstairs; Mrs. come me,'' a with and counter.'' ``your under the ``Then\\
\newline
\textbf{No Proper Nouns}:  ``Then come with me,'' said Mrs. <PROPN>: taking up a dim and dirty lamp, and leading the way upstairs; ``your bed's under the counter.''\\
\newline
\textbf{All Modifications:} taking upstairs <STOP> <STOP> <STOP> leading come lamp way <STOP> <STOP> <PROPN> bed<STOP> <STOP> <STOP> <STOP> dirty counter <STOP> mrs dim said <STOP> <STOP>

\section{Stop Words}
\label{sec:stopwordAppendix}

\subsection{All Stop Words}
\label{sec:allStopWordAppendix}
s, it's, you've, you'll, now, didn't, above, hadn't, has, had, mightn't, don't, for, its, just, she, about, not, his, most, am, we, ll, again, you're, in, aren, or, why, isn't, themselves, you'd, because, the, as, that'll, did, wouldn, couldn't, needn't, to, couldn, a, before, some, been, will, while, re, shouldn, theirs, each, doesn't, isn, be, weren't, any, and, myself, what, hers, all, down, she's, than, our, nor, m, their, these, ve, won't, below, d, it, which, over, how, own, from, shan't, weren, doesn, through, does, if, having, haven, when, too, under, herself, her, wasn't, where, o, ain, itself, mustn't, was, into, they, other, such, those, ours, yourselves, that, himself, them, only, against, this, he, can, very, both, yourself, by, on, hasn, ourselves, more, i, needn, your, won, further, aren't, up, few, then, hadn, with, between, doing, haven't, t, him, an, being, should, there, whom, here, yours, during, shan, didn, so, after, but, wouldn't, do, ma, should've, who, hasn't, mustn, out, is, you, were, have, same, wasn, my, off, once, shouldn't, are, don, y, mightn, me, until, at, no, of

\subsection{Stop Words by Part of Speech}
\label{sec:stopWordPOSAppendix}

\textbf{Adjective:} own, just, other, down, out, not, up, under, same, through, further, few, very, now, only, off, over, in\\
\newline
\textbf{Adverb:} again, between, t, all, as, some, in, then, that, while, when, so, what, no, just, both, nor, this, here, any, before, down, out, not, too, most, up, why, once, but, under, below, same, through, further, there, by, how, each, where, on, very, now, only, above, off, after, about, to, over\\
\newline
\textbf{Conjunction:} as, that, while, when, so, nor, before, for, or, once, but, than, because, where, until, and, now, only, after, if\\
\newline
\textbf{Contraction:} you'd, weren't, mustn, doesn't, it's, wouldn't, hasn, needn, didn, haven, couldn't, needn't, that'll, isn, doesn, mightn't, didn't, hadn, wasn, you've, wouldn, shouldn, don, weren, haven't, you'll, shan, couldn, shan't, aren't, mightn, mustn't, shouldn't, don't, aren, hasn't, isn't, should've, won't, wasn't, you're, she's, hadn't, ain won, re, s, t, d, ll, o\\
\newline
\textbf{Determiner:} some, all, them, an, that, such, more, what, no, which, both, a, his, this, the, any, its, most, our, these, each, few, her, your, their, my, those\\
\newline
\textbf{Noun:} while, no, down, she, up, why, but, ma, m, few, doing, being, have, he, if, all, in, out, do\\
\newline
\textbf{Preposition:} between, from, as, with, in, before, for, down, out, up, during, against, but, under, of, below, through, than, by, until, on, into, at, above, off, after, about, to, over\\
\newline
\textbf{Pronoun:} whom, them, some, me, that, own, such, yourself, you, more, what, which, hers, both, other, his, this, they, any, we, who, most, she, i, himself, themselves, him, these, itself, same, each, few, it, theirs, her, yours, ours, ourselves, myself, herself, those, y, yourselves, he, all\\
\newline
\textbf{Verb:} ve, did, s, own, while, should, re, had, down, out, d, be, up, are, was, is, ll, were, further, been, does, will, can, do, off, has, am, doing, having, being, have, other

\section{Further Results}
\label{sec:results}

\subsection{Accuracy by Class}
\label{sec:accByClass}

\textbf{Samples from In-Training Novels}

\begin{figure}[hbt!]
  \FIG{
      \includegraphics[width=0.4\textwidth, valign=t]{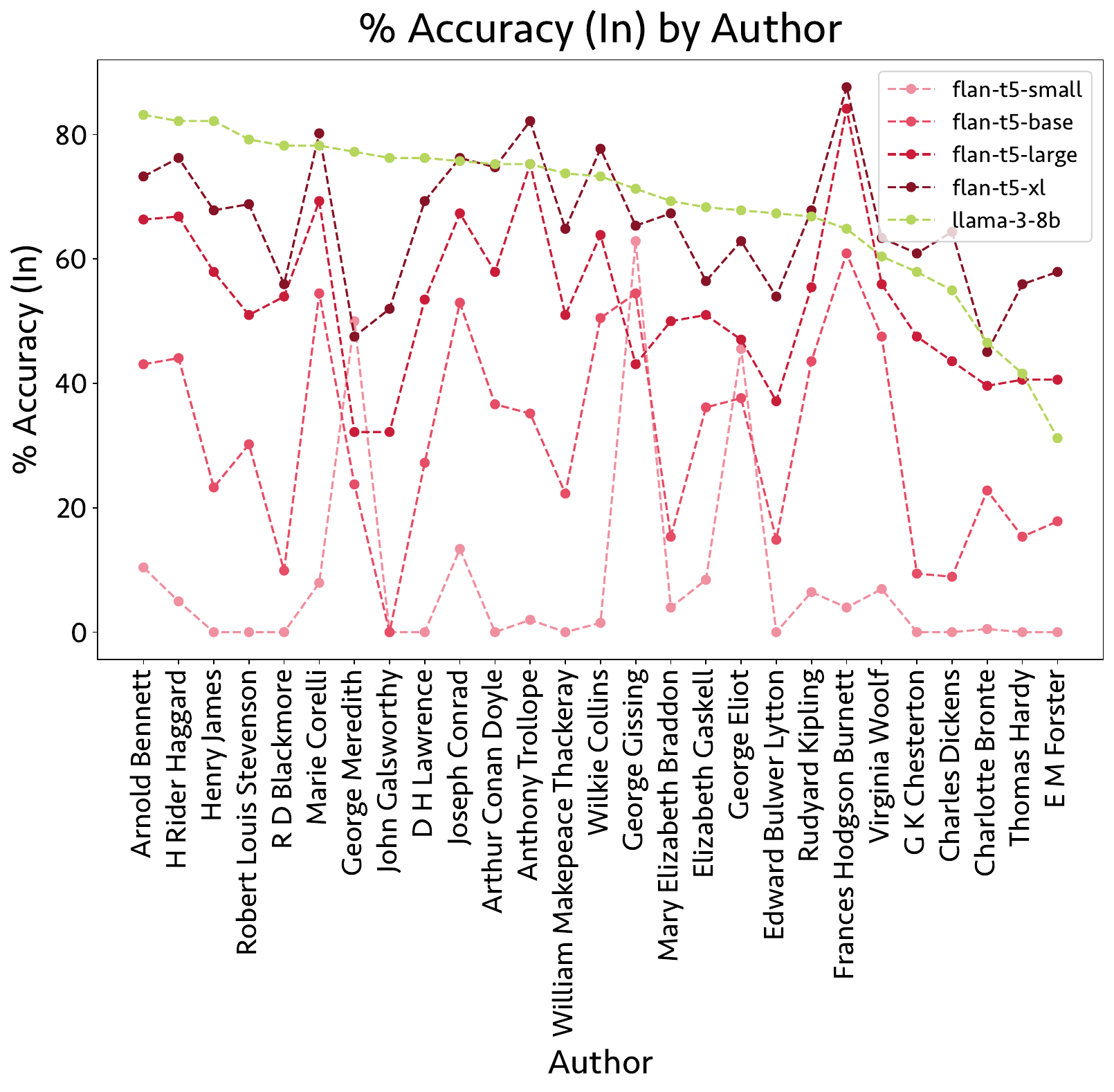} \hspace{6ex}
      \includegraphics[width=0.4\textwidth, valign=t]{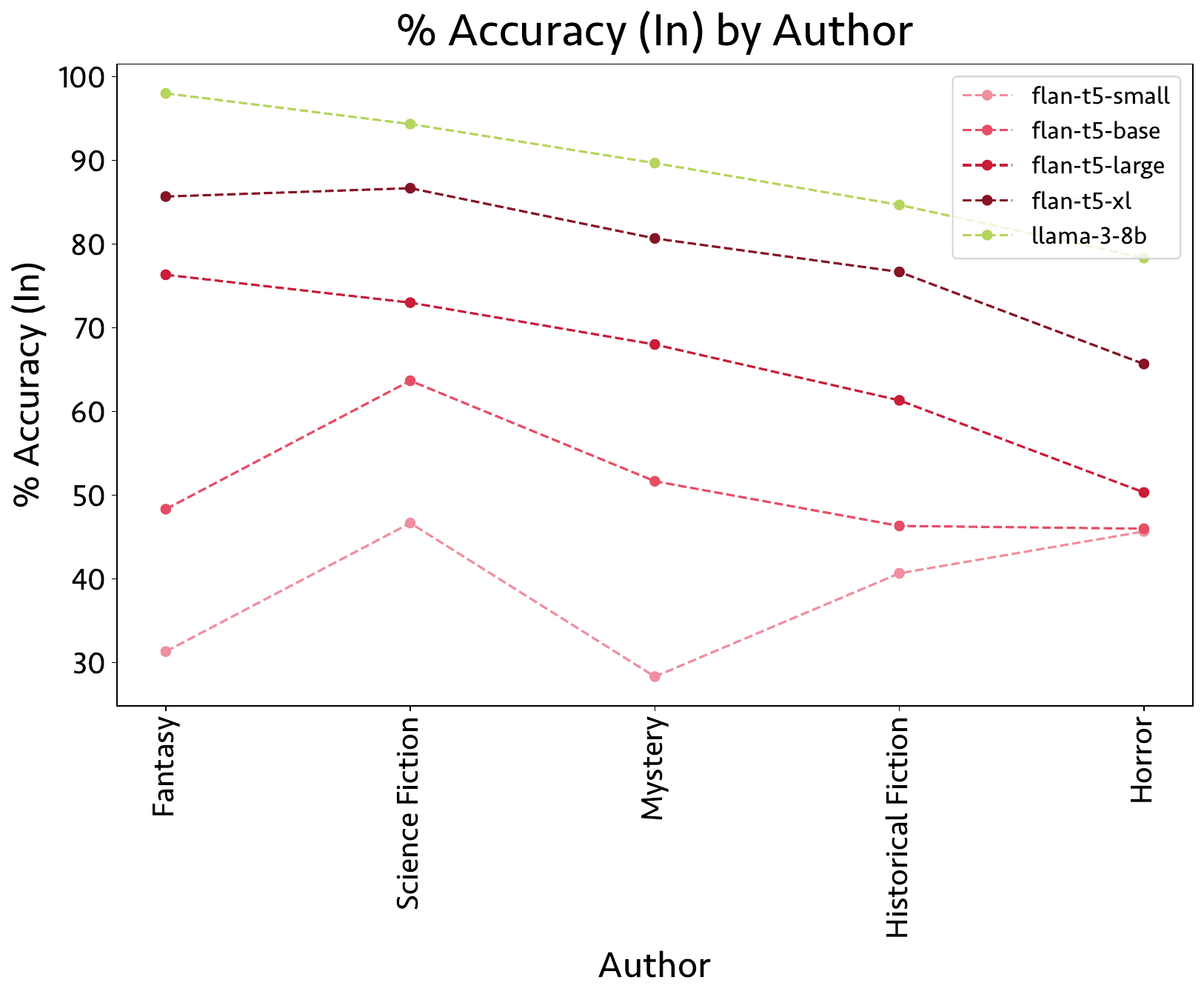}}
  \caption{Accuracy by author (left) and genre (right) for each model on samples from in-training novels. Results of a single run are reported and error bars represent the standard error bootstrapped over 1,000 iterations. The x-axes are sorted by the accuracy of \texttt{llama-3-8b}.}
  \label{fig:accByClassIn}
\end{figure}
\: \newline 
\textbf{Samples from Out-of-Training Novels}

\begin{figure}[hbt!]
  \FIG{
      \includegraphics[width=0.4\textwidth, valign=t]{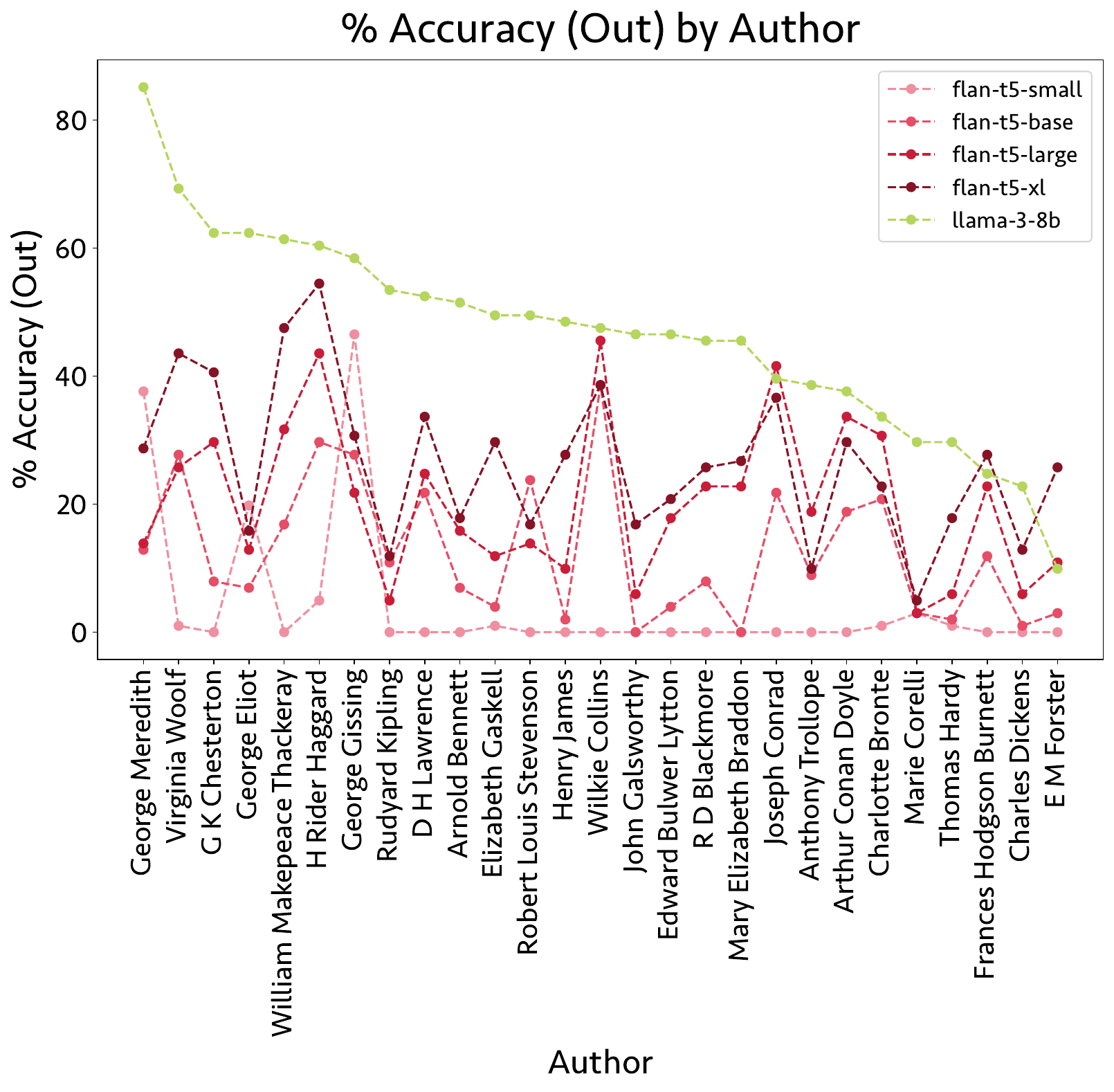} \hspace{6ex}
      \includegraphics[width=0.4\textwidth, valign=t]{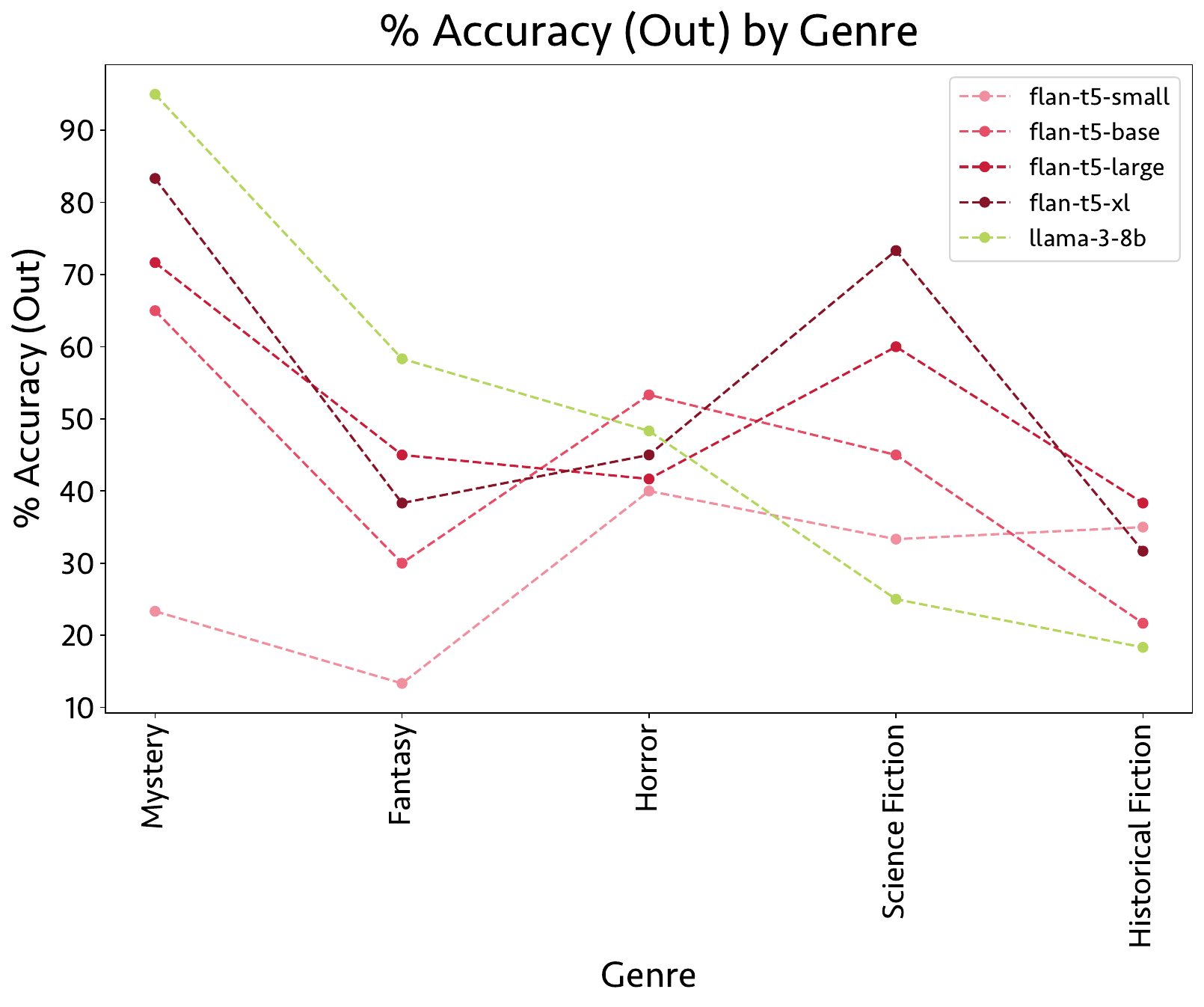}}
  \caption{Accuracy by author (left) and genre (right) for each model on samples from out-of-training novels. Results of a single run are reported and error bars represent the standard error bootstrapped over 1,000 iterations. The x-axes are sorted by the accuracy of \texttt{llama-3-8b}.}
  \label{fig:accByClassOut}
\end{figure}

\subsection{Shapley Values}
\label{sec:shapValues}

\subsubsection{Authorship}
\label{sec:authorShapValues}

\textbf{Top 50 Words by Mean Shapley Values} \\
milliganmy, carmillas, dick, carmilla, bunter, monroe, piggott, grimbold, bunner, walkhams, cosmo, propositions, marquet, trents, ernests, mandersons, deduction, raymond, pringle, freckled, ferrars, thipps, litters, agrippa, gag, grose, staveley, parry, poirot, ferrier, missus, raglan, poirots, ganetts, stangersonwho, mandrake, ackroyd, abbot, defarge, outline, sedan, trent, minas, schloss, bambridge, fleetwood, fordingbridge, catwoman, varney, deathtrap

\: \\
\noindent
\textbf{Top 50 Words by Summed Shapley Values:} \\
the, of, and, to, i, a, in, that, he, was, it, had, his, as, my, with, but, me, for, her, at, him, which, on, she, you, not, we, from, this, there, all, by, they, have, one, be, upon, were, if, is, so, when, little, man, said, out, about, an, been

\subsubsection{Genre}
\label{sec:genreShapValues}

\textbf{Top 50 Words by Mean Shapley Values} \\
brissendens, gibbon, conceited, cormoran, berenice, bullivant, chasten, pigkilling, attenuated, gauchos, rathbone, elliptical, gower, faversham, pearsons, nolan, eunices, sherwoods, bygrave, scurrying, fiance, waltzes, bertie, grandcourts, marillatwas, mavis, enchantment, apemen, fermière, erith, claypole, grandcourt, ledwards, spermlike, corinneshes, deronda, psalmsingers, livia, disciple, wigwam, grimwigs, archery, perfettas, milsom, longshaw, partisans, rosamund, flintcombash, alfredston, unbaptised

\: \\
\noindent
\textbf{Top 50 Words by Summed Shapley Values:} \\
the, a, and, of, to, he, in, that, i, was, her, had, it, his, she, as, with, for, but, you, at, him, if, on, which, not, my, be, all, me, have, this, is, from, so, by, there, said, when, were, one, they, an, would, no, mr, could, been, we, or

\subsection{Attention}
\label{sec:attentionAppendix}

\subsubsection{Authors}
\label{sec:authorAttentionAppendix}

\paragraph{Overall Values}
\label{sec:authorAttentionOverall}
\: \newline 
\textbf{Top 50 Tokens by Mean Attention Value} \\
\_, hav, \_Holmes, \_Alice, <extra\_id\_0>, \_mystery, unter, \_Job, lane, \_THE, ., rot, \_Mormon, \_Harry, \_Dick, \_Jack, )., ]., erson, \_Cosmo, \_|, ca, metry, ?, :, \_Nicholas, \_Martin, worth, \_vampire, \_Tom, \_Arthur, dale, \_William, \_Sherlock, \_Trent, RE, \_Parker, ple, jean, BIT, \_Marius, \_Baker, \_the, \_Echo, ', roy, \_Harris, mill, low, \_Raymond

\: \\
\textbf{Top 50 Tokens by Summed Attention Value} \\
\_, <extra\_id\_0>, \_the, ., :, \_|, RE, \texttt{,}, GEN, a, s, \_of, \_to, ', e, d, \_I, -, \_was, he, \_and, .", \_be, \_it, \_his, t, \_my, ing, \_in, ly, \_an, \_", \_The, ;, \_that, \_you, n, \_me, y, o, \_is, \_him, \_at, \_Alice, ?, \_her, l, \_been, \_were, \_there

\paragraph{Comparing Authors}
\label{sec:authorAttentionAuthors}
\: \newline
\textbf{Arnold Bennett} \\
\hspace*{5mm}\textbf{High:} cie, \_Hall, TH, \_Eve, les, \_|, ella, he, \_Rack, <extra\_id\_0>, OR, \_Prince, sole, ., \_He, \_Violet, \_James, \_Helen, \_the, n \\
\hspace*{5mm}\textbf{Low:} \_, \_I, \_gone, \_my, ;, .", —, \_me, \_Philip, \_", \_are, \_Dick, !", \_Nick, wood, son, \_Lady, \_Lord, rick, in \\
\newline
\textbf{R.\ D.\ Blackmore} \\
\hspace*{5mm}\textbf{High:} \_of, \_Saw, \_Car, AU, \_Major, low, \_Annie, \_Firm, \_For, \_|, \_and, <extra\_id\_0>, :, \_me, ., \_I, \_Do, one, ;, \texttt{,} \\
\hspace*{5mm}\textbf{Low:} \_, -, \_the, \_her, he, \_she, \_She, ', \_He, —, o, \_Philip, t, ?", \_Miss, \_Dick, l, son, !", i \\
\newline
\textbf{Mary Elizabeth Braddon} \\
\hspace*{5mm}\textbf{High:} nton, \_Au, \_Mau, \_her, \_Ellen, son, bank, brook, ley, boy, \_Les, rier, \_Mary, \_Marian, bia, \_Lady, \_Gilbert, \_Robert, \_gone, ' \\
\hspace*{5mm}\textbf{Low:} \_, ., -, —, s, :, y, i, \_Philip, \_Dick, !, AU, \_our, \_their, !", <extra\_id\_0>, \_Helen, \_Nick, \_on, al \\
\newline
\textbf{Charlotte Bronte}\\
\hspace*{5mm}\textbf{High:} e, stone, \_River, \_God, \_Beck, head, \_me, <extra\_id\_0>, \_Graham, \_Paul, :, \_Rochester, \_Caroline, .", —, -, \_Moore, \_my, ;, \_I\\
\hspace*{5mm}\textbf{Low:} ', he, \_had, \_that, \_his, \_Philip, o, \_She, \_But, \_Dick, \_of, \_Sir, \_man, \_Nick, wood, \_Lady, \_him, \_Lord, rick, \_Oliver\\
\newline
\textbf{Frances Hodgson Burnett} \\
\hspace*{5mm}\textbf{High:} \_Er, court, \_Ce, \_Beck, \_Med, roy, \_Hobb, \_and, \_Earl, lock, ric, \_Dick, \_Colin, \_Min, on, \_she, chin, \_Mary, \_Sara, \_ \\
\hspace*{5mm}\textbf{Low:} \texttt{,}, -, \_I, ;, \_the, \_my, \_of, e, ', a, s, \_me, \_is, —, \_we, l, he, ing, \_man, <extra\_id\_0> \\
\newline
\textbf{G.\ K.\ Chesterton} \\
\hspace*{5mm}\textbf{High:} \_Hill, ting, \_or, \_Professor, ?", \_Wilson, \_Valentin, \_Gregory, \_Bull, \_down, er, \_The, \_Father, \_King, \_Bark, \_Wayne, \_Sy, \_Brown, me, \_ \\
\hspace*{5mm}\textbf{Low:} ', \_her, \_gone, \_she, \_I, d, ,, \_my, \_to, \_She, -, \_me, —, e, l, \_Philip, i, t, \_Miss, o \\
\newline
\textbf{Wilkie Collins} \\
\hspace*{5mm}\textbf{High:} \_Margaret, \_Ralph, TH, comb, on, \_Per, \_of, <extra\_id\_0>, hav, -, civ, al, \_Laura, \_to, \_Philip, \_me, \texttt{,}, \_the, \_my, \_I \\
\hspace*{5mm}\textbf{Low:} \_, \_gone, —, \_He, he, n, \_his, \_she, y, \_Dick, son, l, h, er, !", m, \_Nick, wood, \_Lord, \_And \\
\newline
\textbf{Joseph Conrad} \\
\hspace*{5mm}\textbf{High:} al, chi, cou, \_De, \_Ali, he, \_of, \_Alma, \_Da, ul, y, in, \_Nin, \_Pe, \_his, yer, o, a, rol, \_the \\
\hspace*{5mm}\textbf{Low:} \_I, ', ;, \_gone, \_my, \texttt{,}, —, \_you, \_we, n, ., ?, \_Philip, \_me, \_Mr, \_Miss, \_be, \_Dick, son, \_is \\
\newline
\textbf{Marie Corelli} \\
\hspace*{5mm}\textbf{High:} \_Luci, \_are, lini, \_, <extra\_id\_0>, ly, \_Prince, o, \_me, !", \_Zar, \_Robin, noc, \_you, TH, ent, a, !, \_I, - \\
\hspace*{5mm}\textbf{Low:} ;, ,, ., \_He, he, \_that, —, \_his, \_him, \_they, \_was, AU, \_Mr, \_The, \_Philip, \_Dick, son, \_Sir, in, \_Helen \\
\newline
\textbf{Charles Dickens} \\
\hspace*{5mm}\textbf{High:} ley, \_Je, ppy, vis, ham, \_Rose, ick, by, \_me, lock, hav, \_Herbert, \_Leicester, umble, \_Mr, \_I, —, \_, \_Joe, \_Oliver \\
\hspace*{5mm}\textbf{Low:} -, \_she, \_her, \_He, OR, \_of, \_to, \_She, e, l, \_Philip, on, TH, d, \_Dick, <extra\_id\_0>, \_their, \_an, :, \_Helen \\
\newline
\textbf{Arthur Conan Doyle} \\
\hspace*{5mm}\textbf{High:} mouth, \_, \_Professor, \_that, r, s, moor, \_my, leton, \_Sir, TH, \_I, \_is, \_Holmes, \_which, \_us, Saxon, \_our, \_we, \_the \\
\hspace*{5mm}\textbf{Low:} ;, \_her, \_gone, \_she, he, \_He, \_She, \_|, AU, ., —, \_", \_Mr, \_him, \_Philip, ", !, ,", ?", \_Dick \\
\newline
\textbf{George Eliot} \\
\hspace*{5mm}\textbf{High:} .", y, \_Trans, \_her, h, \_H, her, \_Lyon, \_Arthur, en, nah, end, \_gone, \_Harold, ome, ', \_Adam, —, etty, \_Felix \\
\hspace*{5mm}\textbf{Low:} \_the, -, ,, ., \_I, TH, OR, <extra\_id\_0>, !, \_of, ly, \_my, \_|, \_Philip, \_Dick, !", \_Helen, \_Nick, ?", \_me \\
\newline
\textbf{E.\ M.\ Forster} \\
\hspace*{5mm}\textbf{High:} \_Bee, \_Gin, church, \_|, x, \_she, :, erson, \_Abbott, \_Helen, \_Ceci, be, \_Margaret, \_Lili, \_Lucy, AU, .", \_Philip, -, . \\
\hspace*{5mm}\textbf{Low:} \_gone, \_my, ;, \_the, \_I, e, ', a, \_his, s, —, \_, \_me, ly, \_Dick, son, and, \_be, !", \_Nick \\
\newline
\textbf{John Galsworthy} \\
\hspace*{5mm}\textbf{High:} \_James, ren, \_It, ford, \_He, ven, ', \_Din, mes, room, \_Pier, !, \_Fort, ney, \_Clar, \_No, \_, il, \_June, l \\
\hspace*{5mm}\textbf{Low:} \_I, \_my, \_the, \_me, a, d, \_to, —, o, \_Mr, \_Philip, <extra\_id\_0>, \_is, \_Miss, \_we, \_our, \_Dick, al, \_", \_Helen \\
\newline
\textbf{Elizabeth Gaskell} \\
\hspace*{5mm}\textbf{High:} \_Ke, \_Sy, im, \_Brad, \_gone, shaw, ly, borne, \_her, \_Cynthia, son, ster, ', \_, \_Mol, via, ;, \_Ruth, \_Gibson, \_Philip \\
\hspace*{5mm}\textbf{Low:} \_the, a, ., \_I, \_of, s, -, \_my, \_The, —, TH, \_that, <extra\_id\_0>, \_me, \_in, y, e, \_Dick, \_with, al \\
\newline
\textbf{George Gissing} \\
\hspace*{5mm}\textbf{High:} bu, \_artist, foot, \_, \_Virginia, ha, und, \_her, \_War, \_Frank, rton, \_Cross, wood, \_gone, stock, \_Way, \_Monica, \_Julian, \_Will, mark \\
\hspace*{5mm}\textbf{Low:} -, \_the, .", \_my, \_", \_I, —, <extra\_id\_0>, l, \texttt{,}, \_and, s, :, \_me, \_you, ly, \_Philip, OR, ., o \\
\newline
\textbf{H.\ Rider Haggard} \\
\hspace*{5mm}\textbf{High:} \_In, \_of, \_Le, ippo, \_Henry, \_our, i, \_us, yes, s, \_my, \_the, o, \_Hans, \_Good, \_Job, ha, \_gone, \_we, \_I \\
\hspace*{5mm}\textbf{Low:} ', ;, he, \_her, \_He, \_she, \_his, —, \_you, \_Mr, AU, \_him, \_Philip, \_And, \_Miss, \_Dick, ., m, \_Mrs, on \\
\newline
\textbf{Thomas Hardy} \\
\hspace*{5mm}\textbf{High:} ster, !", \_her, \_Joan, s, \_Angel, \_Bol, \_, \_Troy, \_Arab, \_Bath, wood, \_Gabriel, \_Te, —, \_Sue, \_Oak, ella, b, \_Jude \\
\hspace*{5mm}\textbf{Low:} \_I, \_my, n, <extra\_id\_0>, ., ?, \texttt{,}, \_Mr, \_Philip, ;, \_Dick, \_me, on, \_Sir, \_and, \_Helen, \_Nick, o, \_Lady, \_Oliver \\
\newline
\textbf{Henry James} \\
\hspace*{5mm}\textbf{High:} \_Cha, ether, \_Rod, \_her, \_him, \_his, \_Peter, OR, \_Row, \_was, AU, —, ., rick, \_Nick, and, \_He, TH, he, ' \\
\hspace*{5mm}\textbf{Low:} \_, \_my, \_I, !, n, \_did, \_Philip, \_Dick, son, in, \_Sir, \_Helen, o, \_Lord, \_Oliver, \_James, \_father, b, \_Richard, \_me \\
\newline
\textbf{Rudyard Kipling} \\
\hspace*{5mm}\textbf{High:} \_Salt, \_Manuel, \_Tor, pen, \_Di, \_Mai, i, how, —, sko, \_Penn, b, sie, \_Dan, \_Harvey, -, \_Kim, \_Dick, ', \_ \\
\hspace*{5mm}\textbf{Low:} \texttt{,}, \_I, \_her, TH, \_my, \_she, \_the, ;, \_was, \_of, \_She, \_it, \_to, \_me, \_|, \_which, OR, \_had, <extra\_id\_0>, \_been \\
\newline
\textbf{D.\ H.\ Lawrence} \\
\hspace*{5mm}\textbf{High:} \_|, \_Gu, \_George, n, OR, one, \_She, \_And, AU, kin, \_The, etti, \_Ram, :, <extra\_id\_0>, run, \_Kate, \_Gerald, \texttt{,}, . \\
\hspace*{5mm}\textbf{Low:} \_, ;, .", \_my, \_to, \_", —, \_you, \_Mr, \_is, \_be, \_Philip, ?", \_our, \_Miss, \_I, \_Dick, \texttt{,}", son, \_been \\
\newline
\textbf{Edward Bulwer Lytton} \\
\hspace*{5mm}\textbf{High:} uous, \_Chi, \_So, e, \_Da, vers, \_Los, lling, \_Harley, ly, \_Gordon, \_Rand, sper, erton, m, rrell, \_Ken, phy, \_Leonard, \_ \\
\hspace*{5mm}\textbf{Low:} ., \_her, \_the, \_I, ', \_she, \_|, OR, \_was, a, :, \_She, —, \_of, \_we, AU, \_It, \_my, <extra\_id\_0>, \_him \\
\newline
\textbf{George Meredith} \\
\hspace*{5mm}\textbf{High:} \_Tom, \_of, \_Anton, ize, \_Lord, wood, \_my, \_son, \_Austin, !, \_Fleet, \_him, \_Berry, \_Thompson, \_Temple, riot, ', \_father, \_Adrian, \_Richard \\
\hspace*{5mm}\textbf{Low:} .", \_gone, -, \_", OR, —, <extra\_id\_0>, \_she, \_, \_Philip, ?", \_|, \_Dick, l, \_all, ", \_Helen, \_Nick, \_Oliver, \_Mary \\
\newline
\textbf{Robert Louis Stevenson} \\
\hspace*{5mm}\textbf{High:} den, \_we, \_Stewart, less, \_Flint, a, \_Daniel, \_Alan, \_Law, OR, ,, .", \_Silver, \_and, \_me, \_the, \_my, \_Dick, ;, \_I \\
\hspace*{5mm}\textbf{Low:} -, \_she, \_her, ', \_|, \_He, \_, \_which, —, ., \_She, \_an, \_Philip, \_Miss, \_Mrs, \_Helen, \_Nick, !, b, \_Lady \\
\newline
\textbf{William Makepeace Thackeray} \\
\hspace*{5mm}\textbf{High:} un, ?, \_Lord, \_Fo, enden, \_Lambert, \_Wa, \_Major, ker, \_Castle, is, \_Beat, wood, ix, rrington, \_Es, \_Harry, mond, \_Pen, \_ \\
\hspace*{5mm}\textbf{Low:} \_the, a, OR, \_I, \_she, \_|, <extra\_id\_0>, TH, AU, \_her, -, \_to, ., \_was, ,, \_She, \_it, \_of, \_him, — \\
\newline
\textbf{Anthony Trollope} \\
\hspace*{5mm}\textbf{High:} ubb, \_Sir, \_Kennedy, a, \_Violet, \_Bol, hav, \_Mr, \_bishop, \_Lady, \_Colonel, ing, \_Tom, ine, \_Hard, \_, al, \_Ph, \_be, \_gone \\
\hspace*{5mm}\textbf{Low:} ', <extra\_id\_0>, \_I, \texttt{,}, ., \_my, ly, \_me, \_|, \_the, \_and, —, n, AU, OR, :, \_Philip, \_She, TH, \_Dick \\
\newline
\textbf{Virginia Woolf} \\
\hspace*{5mm}\textbf{High:} \_Maggie, \texttt{,}, \_Li, dra, ney, \_Cam, elia, \_they, ?, \_Ram, \_William, \_Mary, say, \_Kath, \_Ralph, \_She, rine, ;, \_her, \_she \\
\hspace*{5mm}\textbf{Low:} \_I, \_my, \_gone, ', \_me, .", \_you, n, —, !, \_is, \_we, i, \_, \_Philip, o, \_Dick, \_our, on, \_Helen \\

\paragraph{Correct vs.\ Incorrect Authors}
\label{sec:authorAttentionCorrect}
\: \newline
\textbf{Correct}: \_loin, \_radiat, direct, AT, \_heal, pyr, \_biography, setting, manager, embracing, \_racing, tip, \_fellowship, max, bot, \_ability, \_duration, deck, aims, \_uses, tube, \_turbulent, duction, colo, \_indication, \_locker, \_separately, \_drip, \_traditional, sheet, \_acres, \_junk, \_threat, one, ripping, \_Marian, reclining, cel, \_Ellen, DA, \_Rod, rine, \_Str, stock, riot, bia, ome, \_Nick, \_Row, rick \\
\newline
\textbf{Incorrect:} \_, \_|, ., \_", \texttt{,}, \_and, ;, AU, \_But, -, ,", \_The, .", \_And, \_He, \_was, \_had, \_When, \_I, :, ', \_that, \_said, \_not, \_It, \_As, he, \_but, \_In, !, \_you, \_There, \_She, I, ?, \_would, \_If, \_were, \_as, \_she, \_have, \_then, \_At, !", \_when, \_His, \_which, —, \_This, ?"

\subsubsection{Genres}
\label{sec:genreAttentionAppendix}

\paragraph{Overall Values}
\label{sec:genreAttentionOverall}
\: \newline
\textbf{Top 50 Tokens by Mean Attention Value} \\
\_, hav, \_Holmes, \_Alice, <extra\_id\_0>, \_mystery, unter, \_Job, lane, \_THE, ., rot, \_Mormon, \_Harry, \_Dick, \_Jack, )., ]., erson, \_Cosmo, \_|, ca, metry, ?, :, \_Nicholas, \_Martin, worth, \_vampire, \_Tom, \_Arthur, dale, \_William, \_Sherlock, \_Trent, RE, \_Parker, ple, jean, BIT, \_Marius, \_Baker, \_the, \_Echo, ', roy, \_Harris, mill, low, \_Raymond \\
\: \\
\textbf{Top 50 Tokens by Summed Attention Value} \\
\_, <extra\_id\_0>, \_the, ., :, \_|, RE, \texttt{,}, GEN, a, s, \_of, \_to, ', e, d, \_I, -, \_was, he, \_and, .", \_be, \_it, \_his, t, \_my, ing, \_in, ly, \_an, \_", \_The, ;, \_that, \_you, n, \_me, y, o, \_is, \_him, \_at, \_Alice, ?, \_her, l, \_been, \_were, \_there

\paragraph{Comparing Genres}
\label{sec:genreAttentionGenre}
\: \\
\textbf{Fantasy} \\
\hspace*{5mm}\textbf{High:} \_|, \_me, !", ic, l, ted, \_they, \_I, f, est, rion, !, \_And, \_she, RE, GEN, <extra\_id\_0>, \texttt{,}, :, \_Alice \\
\hspace*{5mm}\textbf{Low:} \_, ', erson, he, \_his, \_him, d, \_been, \_be, \_Mr, \_an, ation, i, ll, de, le, \_Trent, worth, \_have, rot \\
\newline
\textbf{Historical Fiction} \\
\hspace*{5mm}\textbf{High:} in, )., hos, \_Will, mond, v, \_an, á, \_Nicholas, ry, \_Fred, t, RE, \_Tom, gate, \_King, \_Pierre, he, \_his, \_ \\
\hspace*{5mm}\textbf{Low:} \_I, \_the, \_my, \_Alice, erson, \_me, \_of, \_to, ', i, \_you, \_is, a, \_she, \_we, \_Trent, worth, \_us, rot, \_Peter \\
\newline
\textbf{Horror} \\
\hspace*{5mm}\textbf{High:} on, de, \_there, dale, \_of, \_Jack, ky, \_Arthur, d, \_were, rose, \_horror, \_Ernest, \_Henry, Count, a, \_I, worth, ll, \_my \\
\hspace*{5mm}\textbf{Low:} \_Alice, ?, ., GEN, RE, \_And, c, \_been, \_Trent, <extra\_id\_0>, and, est, rot, \_Peter, ic, rion, \_at, le, \_King, \_mystery \\
\newline
\textbf{Mystery} \\
\hspace*{5mm}\textbf{High:} \_St, \_Holmes, \_case, \_murder, \_Man, low, ang, \_mystery, \_him, le, \_Peter, rot, \_Trent, he, \_it, \_you, \_the, erson, \_to, ' \\
\hspace*{5mm}\textbf{Low:} \_, l, \_Alice, \_were, !, t, \_are, RE, \_their, worth, d, rion, ., ful, ll, est, n, s, ky, \_And \\
\newline
\textbf{Science Fiction} \\
\hspace*{5mm}\textbf{High:} \_its, \_we, \_Dan, ians, \_Circle, -, \_Cost, \_Mart, bra, util, \_Flat, \_Ne, \_our, ., igan, hav, ?, \_Space, and, \_ \\
\hspace*{5mm}\textbf{Low:} <extra\_id\_0>, \texttt{,}, he, \_his, \_to, a, \_Alice, \_him, :, erson, \_was, \_He, RE, \_it, \_", \_|, \_her, \_she, \_the, GEN \\

\paragraph{Correct vs.\ Incorrect Genres}
\label{sec:genreAttentionCorr}
\: \\
\textbf{Correct:} ple, \_fancy, us, \_did, est, rose, -, and, \_you, \_Mart, \_is, f, \_Space, i, \_his, gate, worth, \_horror, c, le, he, ang, m, \_been, \_it, o, \_was, n, y, \_Alice, ly, ing, :, <extra\_id\_0>, l, \_an, erson, t, \_mystery, \_be, RE, \texttt{,}, \_of, d, ', \_to, e, s, a, \_the \\
\newline
\textbf{Incorrect:} \_, \_|, ., .", \_and, \_", ,", \_science, ...", ;, \_As, \_At, \_After, ?", \_but, \_We, \_But, \_ancient, \_old, \_history, \_In, \_--, \_Arthur, \_not, \_for, GEN, \_through, \_curious, \_what, \_Henry, !", \_It, \_This, \_How, \_laboratory, \_dead, \_Still, \_that, \_almost, \_until, \_then, \_Charles, \_how, \_Not, \_The, Besides, \_bat, \_which, \_Time, \_histories

\end{Backmatter}

\end{document}